\documentclass[sigconf,natbib=false,nonacm]{acmart}
\AtBeginDocument{%
  }

\setcopyright{rightsretained}

\RequirePackage[
  datamodel=acmdatamodel,
  style=acmnumeric,
  ]{biblatex}

\usepackage{algpseudocode}
\usepackage{algorithm}
\usepackage{graphicx}
\usepackage{subcaption}  

\usepackage{subcaption}
\usepackage{float} 
\usepackage{multirow}
\usepackage{placeins}
\usepackage{tabularx}
\usepackage{graphicx}
\usepackage{subcaption}

\begin{document}

\title{Sparse Adaptation for In-Context Learning on Data Streams}
\renewcommand{\shorttitle}{Sparse Adaptation for In-Context Learning on Data Streams}


\author{João Neves*}
\orcid{XXXX-XXXX-XXXX-XXXX}
\affiliation{%
  \institution{INESC-TEC, University of Porto}
  \city{Porto}
  \country{Portugal}
}

\author{Afonso Lourenço*}
\orcid{XXXX-XXXX-XXXX-XXXX}
\affiliation{%
  \institution{GECAD, Polytechnic of Porto}
  \city{Porto}
  \country{Portugal}
}

\author{Lizy K. John}
\orcid{0000-0000-0000-0000}
\affiliation{%
  \institution{The University of Texas at Austin}
  \city{Texas}
  \country{United States}
}

\author{Eric P. Xing}
\orcid{XXXX-XXXX-XXXX-XXXX}
\affiliation{%
  \institution{Carnegie Mellon University}
  \city{Pittsburgh}
  \country{United States}
}

\author{João Gama}
\orcid{XXXX-XXXX-XXXX-XXXX}
\affiliation{%
  \institution{INESC-TEC, University of Porto}
  \city{Porto}
  \country{Portugal}
}

\renewcommand{\shortauthors}{João Neves, Afonso Lourenço, Lizy K. John, Eric P. Xing, João Gama}

\begin{abstract}
The Internet of Things (IoT) generates massive data streams, making edge computing essential for low-latency applications and 5G networks. Tabular foundation models (TFMs) enable adaptation through in-context examples rather than parameter updates, but evolving streams challenge context relevance and computational efficiency. This work introduces SAIL, a framework for sparse adaptation of TFMs that reuses the model's key--value cache between adaptation events. SAIL reconstructs its support set only when drifts in stream structure indicate potential context inadequacy, while the structural representation itself is updated by changes in predictive performance. SAIL combines prequential error monitoring with online, class-specific micro-clustering to detect changes in cluster mass, class separability, and feature-space coverage. Upon adaptation, a context budget is allocated proportionally to class mass, and the most recently updated micro-clusters provide representative observations for refitting. This limits context redundancy and computational cost while preserving recent structural information. 
\end{abstract}



\maketitle

\renewcommand{\thefootnote}{\fnsymbol{footnote}}
\footnotetext[1]{These authors contributed equally to this work.}

\section{Introduction}

\begin{figure*}[t]
    \centering
    \includegraphics[width=0.97\textwidth]{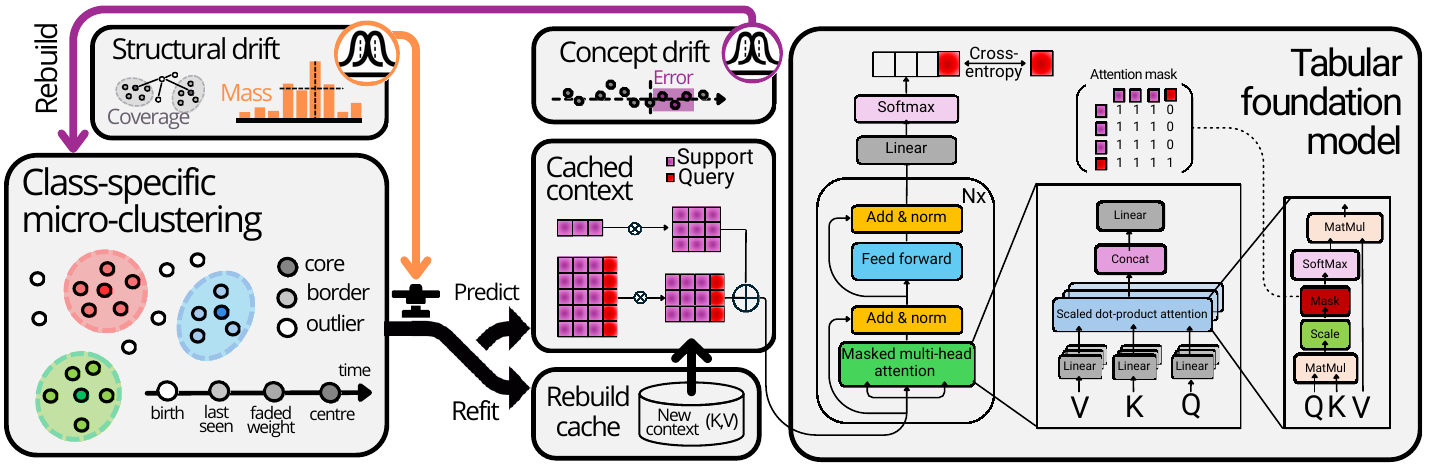}
    \caption{Overview of SAIL. During routine prediction, the TFM reuses its key--value (KV) cache across successive batches. SAIL combines (i) identifies potential context inadequacy by monitoring changes in class-specific micro-cluster mass and feature-space coverage; (ii)  reconstructs contexts, allocating the support-set budget proportionally to class mass and selects the most recently updated micro-clusters; (iii) monitors prequential prediction error and uses backtracking to identify reliable instances for updating the micro-clustering representation. Either drift signal triggers TFM adaptation, with a KV cache rebuilt.}
    \label{fig:vamos}
\end{figure*}

The Internet of Things (IoT) connects vast networks of physical devices, generating massive, high-speed data streams. Edge computing has emerged as a key enabler of low-latency IoT applications and 5G networks, bringing computation closer to data sources and end-users \cite{lourencco2026dfdt,pauperio2026explainable}. However, deploying machine learning at the edge requires not only real-time inference but also continuous adaptation to unbounded, drifting data streams. Tabular foundation models (TFMs) enable adaptation
through in-context examples rather than parameter updates \cite{cho2026causilo,hollmann2025accurate}, but the support set must remain informative as the underlying data distribution evolves, while the cost of context construction and prediction remains compatible with the rate of incoming observations. Although recent TFMs support increasingly large contexts at lower inference times, architectural efficiency alone does not resolve this challenge \cite{lourencco2026context,lourencco2025bridging}. Under evolving streams, accumulating historical observations can increase computational costs without improving predictive performance, particularly when older observations no longer reflect the current predictive relationship \cite{verma2026rethinking}.

\textbf{Sparse adaptation.} Existing approaches to TFM adaptation determine which observations to retain \cite{lee2026bounded,lourencco2025bridging}. Compression methods summarize training observations into learned prompts or latent representations \cite{lin2026context,zabergja2026end}, whereas retrieval-based approaches select examples relevant to individual queries or groups of queries \cite{liu2025tabpfn,ipas2026lucos}. These strategies offer complementary ways to reduce the amount of information processed by the model, including in streaming settings \cite{crista2026streaminghierarchicalinferencetabular,lee2026bounded}. However, such strategies reduce context size without actually reducing adaptation frequency. A compact context may still require frequent reconstruction and refitting, while infrequent updates may leave the model conditioned on an increasingly inadequate support set. A context may resemble the recent distribution while providing insufficient geometric coverage, or span a broad feature space while retaining observations that are no longer representative of the current regime \cite{abdullah2026understandingcontextsamplingtabpfn}. To address this, performance-based drift detectors can identify changes once they become apparent in prediction errors \cite{gama2014survey}, but relevant changes in the stream’s geometric and class structure may arise before they produce a detectable deterioration in prediction quality.

\textbf{This work.} We introduce \textit{SAIL: Sparse Adaptation for In-Context Learning}, a framework for adapting tabular foundation models (TFMs) to evolving data streams. SAIL separates routine prediction from context reconstruction and model fitting, reusing the model's key--value (KV) cache across successive prediction batches. It monitors changes in the relative mass of an online, class-specific micro-clustering to detect structural changes that may compromise context adequacy before they are reflected in predictive error. The micro-clustering representation is maintained incrementally, with decayed cluster weights tracking the evolving distribution of observations. When structural changes trigger adaptation, SAIL constructs a bounded support set by allocating the context budget proportionally to class mass and selecting the most recently updated micro-clusters within each class. This strategy prioritizes recent observations while limiting context redundancy, without modifying the underlying micro-clustering procedure. Separately, ADWIN monitors prequential prediction error. When error rate changes, a backtracking procedure identifies a point from which reliable, stable instances can be used to reconstruct the micro-clustering representation. To validate SAIL, the experimental evaluation characterizes the effects of context size and periodic refitting, compares performance- and structural-based adaptation signals, and evaluates micro-cluster-based context selection, including its computational cost and effects on the evolution of the represented data structure. Code is available at [link after acceptance].

\newpage

\section{Related work}

Tabular foundation models (TFMs) adapt their representational capacity based on in-context data size and structure \cite{cho2026causilo}. Building on this, a new paradigm centers on summarizing unbounded streams on-the-fly before providing them to TFMs \cite{lourencco2025bridging,lourencco2026context}.

\textbf{Data selection.} Existing approaches such as ICD \cite{ma2024context}, TACO \cite{zabergja2026end}, TDColER \cite{kang2025learning}, TL-ANDI \cite{lin2026context}, TFM-Retouche \cite{nguyen2026tfm}, and TuneTables \cite{feuer2024tunetables} compress training data into learned prompts or latent prototypes. Other methods perform hard data selection, with iterative sample and feature pruning using cached attention statistics \cite{rundel2024interpretable}, Kernel SHAP-inspired utility estimation \cite{chen2026vip}, or ridge regression surrogates of influence \cite{benihaddadene2026ticda}. Several works argue contexts should be retrieved to each query to not only improve performance, but also reduce inference cost \cite{mohamed2026steer,ning2026test}. LoCalPFN \cite{thomas2024retrieval}, TabDPT \cite{ma2024tabdpt}, In-Context \cite{nejjar2024context}, and BETA \cite{liu2025tabpfn} retrieve neighbors in the input space, whereas LETabPFN \cite{koshil2024towards} and LUCOS \cite{ipas2026lucos} operate in learned representation spaces. For streaming scenarios, HINT \cite{crista2026streaminghierarchicalinferencetabular} maintains an approximate nearest-neighbor graph over sliding windows. Alternatively, MICP \cite{xu2024mixture} and CRUMB \cite{heredge2026crumb} amortize retrieval by clustering similar queries to share a common support set, recovering efficient batching without sacrificing localized learning. QCOC \cite{zhao2026compile} directly clusters joint key--value states into compact prototypes.

\textbf{Resource constraints.} These data selection strategies are designed for static datasets or settings in which constant batch updates are affordable. Under evolving streams, however, repeatedly evaluating instance importance or reconstructing contexts can impose substantial computational overhead \cite{verma2026rethinking,lourencco2025device}. This overhead matters because an adaptation strategy must process data at a rate compatible with the stream's arrival rate. Otherwise, computationally intensive methods may fail to keep pace with incoming observations, undermining the validity of performance comparisons in high-velocity settings \cite{ghunaim2023real,alfarra2023evaluation}. Recent work has therefore emphasized computational constraints and time-dependent utility, including deadline-, latency-, and budget-aware evaluation \cite{prabhu2023computationally,sreeram2026tempora}. Streaming TFM systems must consequently coordinate replay and update scheduling \cite{liu2025latency}, exploit representative memories to support sparse updates \cite{cha2026snap,risca2025boosting}, and account for both data arrival rates \cite{zhang2025one} and label latency \cite{souza2020evaluating}. These considerations motivate \emph{sparse adaptation}: selectively updating the model's in-context data rather than continuously reconstructing or expanding its context.

\section{Preliminaries}
\label{sec:preliminaries}

\begin{figure}[t]
    \centering
    \begin{subfigure}{0.23\textwidth}
        \centering
        \includegraphics[width=\linewidth]{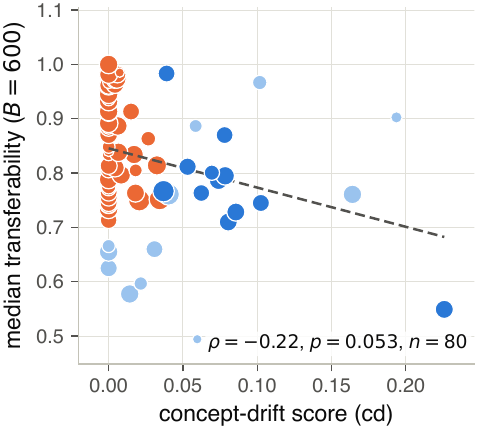}
        \caption{Transfer vs. concept drift}
        \label{fig:first}
    \end{subfigure}
    \hfill
    \begin{subfigure}{0.23\textwidth}
        \centering
        \includegraphics[width=\linewidth]{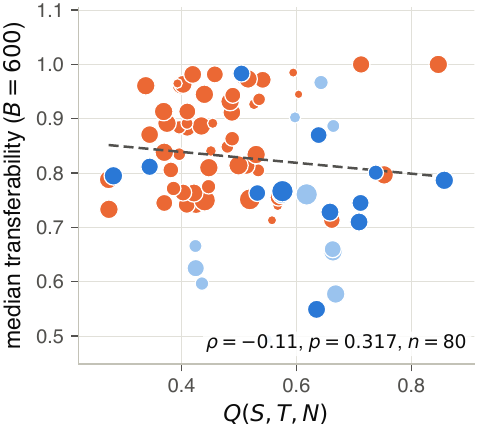}
        \caption{Transfer vs. temporal changes}
        \label{fig:second}
    \end{subfigure}

    \par\medskip
    \begin{subfigure}{0.24\textwidth}
        \centering
        \includegraphics[width=\linewidth]{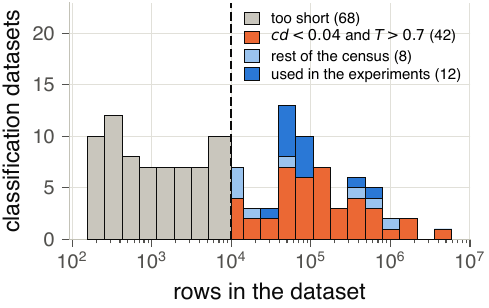}
        \caption{Dataset size}
        \label{fig:third}
    \end{subfigure}
    \hfill
    \begin{subfigure}{0.23\textwidth}
        \centering
        \includegraphics[width=\linewidth]{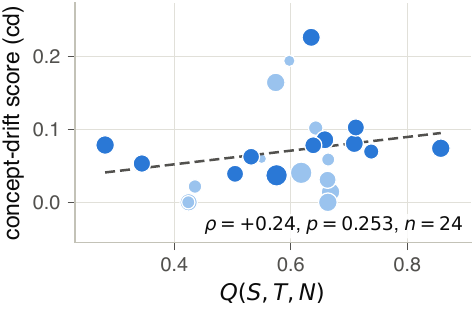}
        \caption{Concept drift vs. temporal}
        \label{fig:fourth}
    \end{subfigure}
    \caption{Streams selected across 80 benchmarks.}
    \label{fig:vamos}
\end{figure}

Recent work shows that larger contexts do not necessarily improve performance, with a smaller sliding-window approach outperforming on an abrupt-drift stream \cite{verma2026rethinking}. Conversely, larger contexts can increase diversity and feature-space coverage, both of which are important for effective contexts \cite{abdullah2026understandingcontextsamplingtabpfn}. Together, these findings highlight that the effectiveness of sparse adaptation depends not only on distributional alignment but also on how well retained examples cover the evolving feature space. We therefore explicitly seek challenging streams in which these properties vary over time. To this end, we collect 80 benchmarks from six repositories: USP \cite{souza2020challenges}, OEBench \cite{diao2023oebench}, TabRED \cite{rubachev2024tabred}, BeyondArena \cite{purucker2026beyond}, TableShift \cite{gardner2023benchmarking}, and WhyShift \cite{liu2023need}. Each stream is divided into consecutive, non-overlapping batches
$\mathcal{Q}_t$ of
${300,600,1000}$ instances. We characterize transitions using feature-mean discrepancy ($S$), class-distribution discrepancy ($T$), and coverage gap ($N$):
\begin{table}[t]
\centering
\caption{Twelve selected streams.}
\label{tab:streams}
\setlength{\tabcolsep}{2pt}
\small
\renewcommand{\arraystretch}{0.8}
\begin{tabular}{lrrr|rrrr|rr}
\toprule
Stream & $n$ & $d$ & $C$ & $S$ & $T$ & $N$ & $Q$ & \footnotesize transf. & $cd$ \\
\midrule
{\footnotesize rialto}\textsuperscript{1,3} & 82\,250 & 27 & 10 & .297 & .927 & .681 & .635 & .549 & .226 \\
{\footnotesize ins\_inc\_abrupt\_bal}\textsuperscript{1,3} & 79\,986 & 33 & 6 & .703 & .759 & .664 & .709 & .710 & .080 \\
{\footnotesize ins\_inc\_reoc\_bal}\textsuperscript{1,3} & 79\,986 & 33 & 6 & .709 & .734 & .531 & .658 & .728 & .086 \\
{\footnotesize ins\_abrupt\_bal}\textsuperscript{1,3} & 52\,848 & 33 & 6 & .825 & .762 & .547 & .711 & .745 & .102 \\
{\footnotesize electricity}\textsuperscript{1} & 45\,312 & 8 & 2 & .653 & .402 & .541 & .532 & .763 & .062 \\
{\footnotesize ins\_inc\_reoc\_imbal}\textsuperscript{1,3} & 452\,044 & 33 & 6 & .669 & .784 & .273 & .576 & .767 & .037 \\
{\footnotesize covertype}\textsuperscript{3,4} & 100\,000 & 54 & 7 & .925 & .725 & .922 & .857 & .787 & .074 \\
{\footnotesize rbf}\textsuperscript{5} & 100\,000 & 10 & 5 & .206 & .252 & .386 & .281 & .795 & .078 \\
{\footnotesize ins\_gradual\_bal}\textsuperscript{1,3} & 24\,150 & 33 & 6 & .856 & .833 & .525 & .738 & .801 & .069 \\
{\footnotesize tree\_recurring}\textsuperscript{5} & 60\,000 & 10 & 2 & .281 & .481 & .270 & .344 & .812 & .053 \\
{\footnotesize us\_accident}\textsuperscript{2} & 49\,000 & 30 & 2 & .588 & .787 & .541 & .639 & .870 & .078 \\
{\footnotesize agrawal}\textsuperscript{5} & 50\,000 & 9 & 2 & .439 & .614 & .459 & .504 & .983 & .039 \\
\bottomrule
\end{tabular}
\vspace{2pt}
\begin{minipage}{\linewidth}
\footnotesize
\textsuperscript{1} USP \cite{souza2020challenges}
\textsuperscript{2} WhyShift \cite{liu2023need}
\textsuperscript{3} OEBench \cite{diao2023oebench}
\textsuperscript{4} BeyondArena \cite{purucker2026beyond}
\textsuperscript{5} Synthetic
\end{minipage}
\end{table}
\begin{align}
S_t &=
\frac{1}{d}\sum_{j=1}^{d}
\left|\mu_j(\mathcal{Q}_t)-\mu_j(\mathcal{Q}_{t-1})\right|,
\\
T_t &= \frac{1}{2}\sum_{c=1}^{K}
\left|p_t(c)-p_{t-1}(c)\right|,
\\
N_t &= \frac{1}{B}\sum_{x\in\mathcal{Q}_t}
\min_{z\in\mathcal{Q}_{t-1}}\|x-z\|_2.
\end{align}
where $\mu_j(\mathcal{Q}_t)$ is the mean of feature $j$ and $p_t(c)$ is the proportion of class $c$ in batch $\mathcal{Q}_t$. To account for dataset-specific magnitudes, descriptor trajectories are standardized within each stream using robust location and scale estimates. For descriptor $q$, its relative value at time $t$ is
[
$z_t^{(q)}=
\frac{q_t-\operatorname{median}(q)}
{Q_{0.75}(q)-Q_{0.25}(q)}$
],
where $Q_p(q)$ denotes the empirical $p$-quantile. Descriptors with zero interquartile range are excluded from this standardization. Four complementary characteristics capture their temporal behaviour: relative excursion magnitude ($E_{95}(q)$), elevated-state prevalence ($P_{\kappa}(q)$), temporal intermittency ($I_{\kappa}(q)$), and relative change concentration ($C_{10}(q)$). First, \textit{relative excursion magnitude} \(E_{95}(q)\) characterizes the upper range of the descriptor trajectory through the 95th percentiles of $z_t^{(q)}$. Second, \textit{elevated-state prevalence} measures the proportion of
the trajectory spent above a fixed dimensionless threshold $\kappa$:
\begin{equation}
P_{\kappa}(q)
=
\frac{1}{m}\sum_{t=1}^{m}
\mathbb{1}\!\left[z_t^{(q)}>\kappa\right].
\end{equation}
Third, \textit{temporal intermittency} characterizes the variability
in the durations of elevated-value episodes. An episode is defined
as a maximal sequence of consecutive observations satisfying
$z_t^{(q)}>\kappa$. Let $L_1,\ldots,L_J$ denote the durations of the
$J$ identified episodes. Intermittency is quantified by their
coefficient of variation:
\begin{equation}
I_{\kappa}(q)
=
\frac{\operatorname{sd}(L_1,\ldots,L_J)}
{\operatorname{mean}(L_1,\ldots,L_J)}.
\end{equation}
Finally, \textit{relative change concentration} characterizes whether
temporal variation is distributed across many transitions or
concentrated in a few large changes. Let $\mathcal{T}_{10}$ contain the largest
$\lceil 0.1(m-1)\rceil$ transition changes $r_t^{(q)}$, the change concentration is defined as:
\begin{equation}
C_{10}(q)=
\frac{\sum_{t\in\mathcal{T}_{10}}r_t^{(q)}}
{\sum_{t=2}^{m}r_t^{(q)}}, \quad r_t^{(q)} =
\left|z_t^{(q)}-z_{t-1}^{(q)}\right|,
\quad t=2,\ldots,m,.
\end{equation}
Together, these four characteristics capture the magnitude, prevalence, intermittency, and temporal concentration of changes in each descriptor trajectory. They are computed separately for $S$, $T$, and $N$, yielding twelve descriptors in total. The aggregate score $Q$ is computed as the equally weighted mean of the 12 normalized descriptors. Beyond temporal descriptors, transferability is assessed by evaluating a TFM in-context on the preceding batch. Concept drift is measured using a scikit-learn decision tree, detecting accuracy drops between consecutive batches that exceed the shuffled-stream baseline by three standard deviations; the final score combines drift frequency and severity. Figure \ref{fig:first} shows their relation. Figures~\ref{fig:second} and~\ref{fig:fourth} show the relationship of $Q$ with predictive transferability and temporal separability, respectively, while Figure~\ref{fig:third} summarizes dataset sizes. Guided by these profiles, we first retain streams with concept drift above $0.04$, or predictive transferability below $0.7$. Among the remaining candidates, we select 12 streams, with at least 10,000 instances, that provide broad coverage of temporal behaviours across the twelve descriptors. Table~\ref{tab:streams} summarizes the selected streams, with additional details on the non-selected datasets in Appendix~\ref{sec:results}.

\section{Methodology}

Let a labeled data stream be denoted by
$\mathcal{S}=\{(\mathbf{x}_t,y_t)\}_{t=1}^{\infty}$,
where $\mathbf{x}_t\in\mathbb{R}^{M}$ is an $M$-dimensional feature vector and $y_t\in\mathcal{Y}$ is a finite set of classes. The joint distribution of features and targets may change over time. Given a labeled support set $\mathcal{D}_t^{\mathrm{ctx}}$ and a batch of query instances $\mathbf{X}_t^q$, a pretrained TFM with parameters $\theta$ produces predictions
$\widehat{\mathbf{Y}}_t
=
f_{\theta}\left(
\mathbf{X}_t^q,
\mathcal{D}_t^{\mathrm{ctx}}
\right).$
Unlike conventional streaming approaches that update the predictor continuously, SAIL keeps the support set and its associated KV cache fixed between adaptation events. Predictions are produced in batches of $100$ queries, while incoming observations are processed individually to update streaming statistics, drift detectors, and memory structures. These updates do not require refitting the TFM during ordinary prediction. SAIL comprises three complementary mechanisms: (i) structural drift detection, which tracks the evolution of class-specific micro-clusters and identifies changes that may precede measurable predictive degradation;
(ii) context compression, which allocates the support-set budget proportionally to class mass and selects the
most recently updated micro-clusters; and (iii) performance-based drift detection, which monitors the cached TFM prediction error and estimates the onset of detected changes in order to update the micro-clustering itself. Either drift-detection mechanism can trigger context reconstruction and subsequent refitting.

\subsection{Structural drift}

Performance-based detection responds to changes reflected in predictive error, but changes in the stream's structure may arise before they produce a detectable deterioration in prediction quality. SAIL therefore monitors an online, class-specific micro-cluster representation. Micro-clustering extends the cluster-feature summaries introduced in BIRCH \cite{zhang1996birch} and their streaming formulation in CluStream \cite{han2003framework}. Each micro-cluster summarizes a group of observations through its accumulated statistics, centroid, weight, and temporal information. For each class $c\in\mathcal{Y}$, let
$\mathcal{M}_c^{(t)}$ denote the micro-clusters at time $t$. A micro-cluster $\mu_k$ has centroid $\mathbf{z}_k$, decayed weight $w_k^{(t)}$, and characteristic radius $r_k$. Upon observing a labeled instance $(\mathbf{x}_t,y_t)$ with $y_t=c$, the corresponding class-specific structure is updated. Let
\begin{equation}
\mathcal{N}_{\varepsilon}(\mathbf{x}_t)
=
\left\{
\mu_k\in\mathcal{M}_c^{(t)}:
\|\mathbf{x}_t-\mathbf{z}_k\|_2
\leq\varepsilon
\right\}.
\end{equation}
For each micro-cluster in this neighborhood, its weight and centroid are updated as
\begin{equation}
w_k\leftarrow w_k+1,
\qquad
\mathbf{z}_k\leftarrow
\mathbf{z}_k+
\frac{1}{w_k}(\mathbf{x}_t-\mathbf{z}_k).
\end{equation}
If a neighborhood is empty, a new micro-cluster is created at $\mathbf{x}_t$. To accommodate gradual changes, cluster weights decay according to
\begin{equation}
w_k^{(t)}=w_k^{(t')}\,2^{-\lambda\,(t-t')},\qquad \lambda>0,
\label{eq:fade}
\end{equation}
where $\Delta t$ is the elapsed time since the preceding update. Micro-clusters whose weights fall below a threshold $\tau$ are removed.

\textbf{Cached context inadequacy.} The online micro-cluster representation summarizes the evolving stream, while the representation at the previous fit characterizes the stream when the cached TFM context was constructed. SAIL monitors \textit{mass discrepancy} to quantify changes in the relative distribution of micro-cluster weights between these two representations. Let $W_c^{(t)}=\sum_{\mu_k\in\mathcal{M}_c^{(t)}}w_k^{(t)}$ and $W^{(t)}=\sum_c W_c^{(t)}$. Define the normalized within-class weight of micro-cluster $\mu_k$ as
\[
p_k^{(t)}=\frac{w_k^{(t)}}{W_c^{(t)}}.
\]
The change in mass distribution between the previous fit at $t_0$ and the current time $t$ is measured by
\begin{equation}
\begin{aligned}
\Delta_{\mathrm{mass}}
&=
\sum_{c\in\mathcal{Y}}
\frac{W_c^{(t)}}{W^{(t)}}\,
\delta_c(t_0,t),\quad
\delta_c(t_0,t)
=
\begin{cases}
1, &
\substack{
\text{if } W_c^{(t_0)}=0\\
\text{or } W_c^{(t)}=0,
}\\[6pt]
\frac{1}{2}D_c(t_0,t),
& \text{otherwise},
\end{cases}\\[4pt]
D_c(t_0,t)
&=
\sum_{k\in\mathcal{U}_c}
\left|p_k^{(t)}-p_k^{(t_0)}\right|,\quad
\mathcal{U}_c =
\mathcal{M}_c^{(t_0)}
\cup
\mathcal{M}_c^{(t)}.
\end{aligned}
\label{eq:mass}
\end{equation}
Micro-cluster identifiers are never reused after deletion, preventing distinct micro-clusters from being incorrectly matched across snapshots and the mass change from being underestimated. A cluster absent at either time is assigned zero weight. Thus, cluster births, deaths, and changes in relative mass contribute to the statistic. The computation is linear in the number of aligned micro-clusters. Structural statistics are evaluated relative to their values at the previous fit, and a structural alarm is raised when their changes exceed the prescribed threshold $\theta$. Such alarms trigger context reconstruction, and cache rebuilding. Unlike a performance-based alarm, a structural alarm does not reset the performance-based drift detector or rebuild the underlying online micro-clustering structure.

\begin{figure*}[t]
    \centering
    \begin{subfigure}{0.24\textwidth}
        \includegraphics[width=\linewidth]{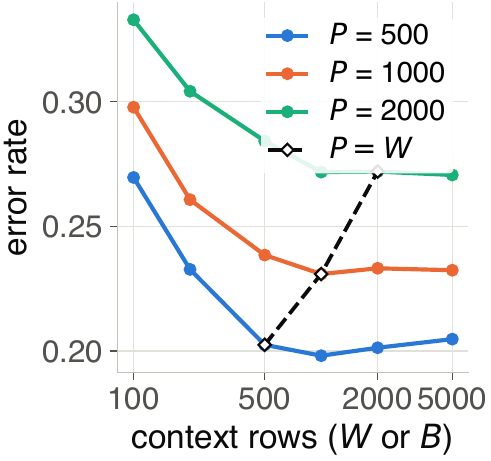}
        \caption{Error rate}
        \label{fig:h1_a}
    \end{subfigure}
    \begin{subfigure}{0.24\textwidth}
        \includegraphics[width=\linewidth]{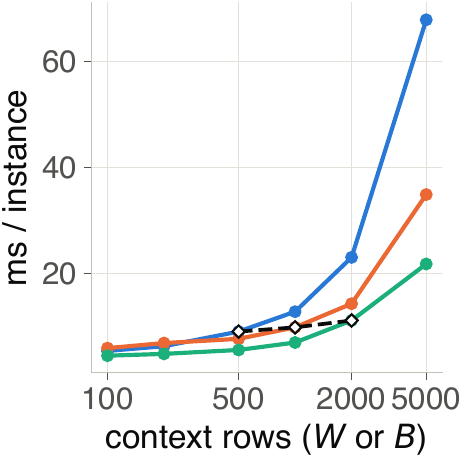}
        \caption{Inference cost}
        \label{fig:h1_b}
    \end{subfigure}
    \begin{subfigure}{0.24\textwidth}
        \includegraphics[width=\linewidth]{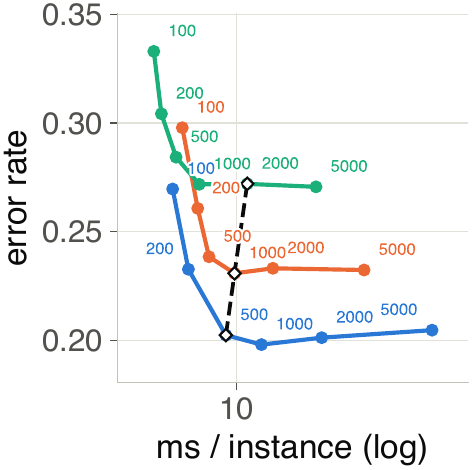}
        \caption{Error-cost trade-off}
        \label{fig:h1_c}
    \end{subfigure}
    \begin{subfigure}{0.24\textwidth}
        \includegraphics[width=\linewidth]{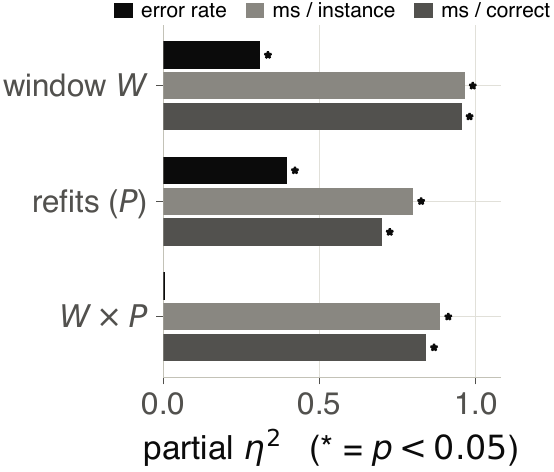}
        \caption{ANOVA effects}
        \label{fig:h1_d}
    \end{subfigure}
    \caption{Effect of periodically updated sliding-window contexts on predictive performance and computational cost.}
    \label{fig:h1}
\end{figure*}

\subsection{Context compression}

Between adaptation events, prediction batches reuse the existing KV cache. When an adaptation event occurs, observations collected since the estimated change point form the candidate support set for refitting the TFM, after which the KV cache is rebuilt. However, fitting on every observation can be unnecessarily expensive when many observations occupy similar regions of the feature space. SAIL therefore reuses the micro-cluster representation maintained for structural change detection to construct a less redundant support set for refitting. Using one representative per active micro-cluster already reduces the support set relative to using all observations. When a fixed context budget $B$ is required, the number of representatives allocated to each active class $c$ is determined by its share of the total decayed mass:
\[
b_c=\min\left\{|\mathcal{M}_c^{(t)}|,
\max\left(1,\operatorname{round}\left(
B\frac{W_c^{(t)}}{\sum_{c'}W_{c'}^{(t)}}
\right)\right)\right\},
\]
where $W_c^{(t)}=\sum_{\mu_k\in\mathcal{M}_c^{(t)}}w_k^{(t)}$. Within each class, the $b_c$ most recently updated micro-clusters are selected, each represented by its latest real observation. This selection is recomputed at every refit, allowing previously excluded clusters to re-enter the context as the stream evolves and they receive new observations. The budget therefore controls context size through representative selection, without modifying the underlying micro-cluster maintenance or drift detection, enabling the effect of pruning older representatives to be evaluated independently.

\subsection{Performance drift}

Structural changes can indicate that the cached context no longer adequately represents the evolving stream, but they do not establish whether its continued use degrades predictive performance. Conversely, predictive deterioration may occur even when changes in the stream's local geometry are subtle. SAIL therefore complements structural monitoring with an independent performance-driven forgetting mechanism.

\textbf{Prequential error monitoring.} Let $\widehat{y}_t^{\mathrm{TFM}}$ denote the prediction produced by the tabular foundation model (TFM), and define the prequential classification error as
$e_t=\mathbb{1}[\widehat{y}_t^{\mathrm{TFM}}\neq y_t]$.
ADWIN \cite{bifet2007learning} monitors this error sequence and raises an alarm when its adaptive window detects a statistically significant distributional change. The alarm signals a change in predictive performance but does not directly identify the onset of the new regime. SAIL therefore backtracks from the alarm to estimate a candidate change point, which defines the boundary for context reconstruction. Observations following this boundary are retained to construct the replacement TFM context. To ensure that the resulting support set is sufficiently large and diverse for classification, the reconstruction interval is expanded whenever it contains fewer than 200 observations. The retained observations replace the context, and the cache is rebuilt.

\section{Experimental evaluation}
\label{sec:experiments}

The experimental evaluation is organized around four research questions that progressively motivate the design of SAIL. First, how do periodically updated sliding-window contexts affect feature-mean discrepancy, class-distribution discrepancy, and coverage gap, as well as trade predictive performance against computational cost under concept drift? (RQ1) Second, can micro-cluster-based compression reduce context redundancy and computational cost without substantially degrading prediction quality? (RQ2) Third, do mass discrepancy signals improve computation effort where it matters, capturing and affecting structural prediction errors in relevant ways? (RQ3) Fourth, does the proposed performance-based and mass discrepancy signals outperform alternative adaptive or fixed periodic policies? (RQ4)

\begin{figure}[t]
    \centering
    \begin{subfigure}{0.32\linewidth}
        \includegraphics[width=\linewidth]{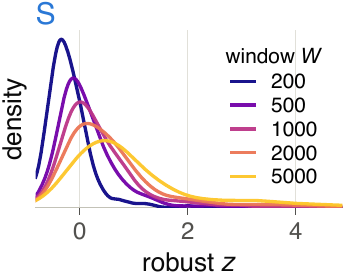}
        \caption{$\Delta$ feature-mean }
    \end{subfigure}
    \begin{subfigure}{0.33\linewidth}
        \includegraphics[width=\linewidth]{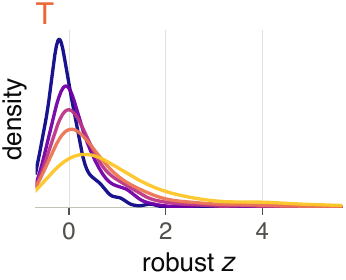}
        \caption{$\Delta$ class-distribution}
    \end{subfigure}
    \begin{subfigure}{0.32\linewidth}
        \includegraphics[width=\linewidth]{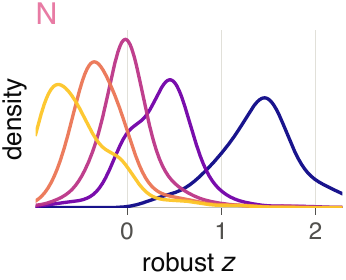}
        \caption{Coverage gap}
    \end{subfigure}

    \par\medskip

    \begin{subfigure}{0.32\linewidth}
        \includegraphics[width=\linewidth]{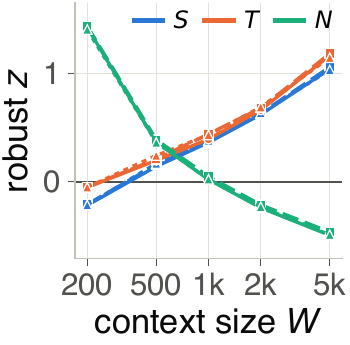}
        \caption{Mean}
        \label{fig:juju_d}
    \end{subfigure}
    \begin{subfigure}{0.32\linewidth}
        \includegraphics[width=\linewidth]{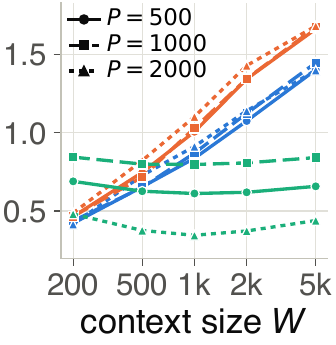}
        \caption{Std. deviation}
        \label{fig:juju_e}
    \end{subfigure}
    \begin{subfigure}{0.32\linewidth}
        \includegraphics[width=\linewidth]{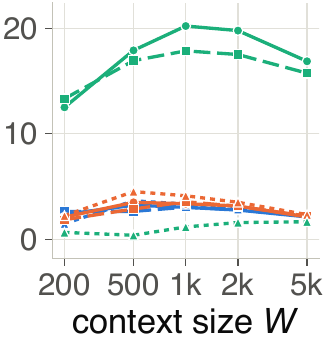}
        \caption{Skewness}
        \label{fig:juju_f}
    \end{subfigure}

    \caption{Effect of periodic sliding-window contexts on $\Delta$ feature-mean, $\Delta$ class-distribution, and coverage gap.}
    \label{fig:juju}
\end{figure}

\begin{figure*}[t]
    \centering
    \includegraphics[width=\textwidth]{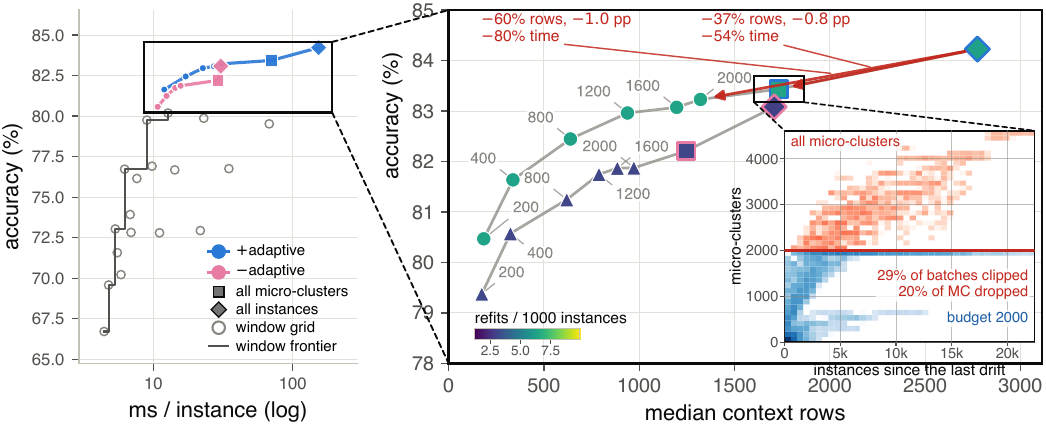}
    \caption{Effect of SAIL's class-specific micro-cluster compression on inference cost, context size, and accuracy loss.}
    \label{fig:compression}
\end{figure*}

\textbf{More context does not necessarily improve performance (RQ1).}
Under non-stationary conditions, increasing the amount of historical data in a TFM context does not guarantee better predictive performance. Figure~\ref{fig:h1} evaluates sliding-window sizes $W\in\{100,300,500,1500,2000,5000\}$ and periodic refitting intervals $P\in\{500,1000,2000\}$. Figure~\ref{fig:h1_a} shows that increasing context size initially reduces the error rate, but the gains diminish and can eventually reverse as older observations become less representative of the current stream. More frequent refitting generally reduces error, while no substantial interaction between window size and refitting frequency is observed within the evaluated range. Figure~\ref{fig:h1_b} shows that inference cost grows increasingly steeply with context size, particularly at higher refitting frequencies. Figure~\ref{fig:h1_c} consequently highlights the trade-off between predictive performance and inference cost. This is further supported by Figure~\ref{fig:h1_d}: refitting frequency has a stronger effect on error rate than context size, with little interaction between the two factors, whereas inference cost is more sensitive to context size and exhibits an interaction with refitting frequency. Figures~\ref{fig:juju_d}--\ref{fig:juju_f} further characterize the trade-offs between feature-mean discrepancy, class-distribution discrepancy, and coverage gap. Figure~\ref{fig:juju_d} shows that larger windows reduce the mean coverage gap but increase both the mean feature-mean discrepancy and the mean class-distribution discrepancy. Figure~\ref{fig:juju_e} shows that the standard deviation of the coverage gap remains relatively stable across window sizes, whereas the variability of both distributional discrepancies increases with larger windows. Thus, the benefits of smaller contexts in limiting distributional discrepancies depend on dataset-specific characteristics. Finally, Figure~\ref{fig:juju_f} shows pronounced skewness in the coverage gap at longer refitting intervals, suggesting that delayed context updates can lead to occasional episodes of particularly poor coverage. Together, these findings motivate adaptive context selection.

\textbf{Micro-cluster compression reduces inference cost with limited accuracy loss (RQ2).} Fitting a TFM on every observation can be unnecessarily expensive when many observations represent similar regions of the feature space. Figure \ref{fig:compression} evaluates the trade-off between context size, inference cost, and predictive accuracy under two adaptation sensitivities. $\Delta_{\text{mass}}^{+adapt}$ uses an ADWIN significance level of 0.02 and a structural threshold of 0.05, whereas $\Delta_{\text{mass}}^{-adapt}$ uses 0.1 and 0.02, respectively. Both configurations use a clustering radius of 0.5, a cleanup interval of 2000, a warmup of 500, and a fading factor of $10^{-4}$. These settings induce different adaptation frequencies and, consequently, different accuracy--cost trade-offs across context budgets. For instance, using all instances, $\Delta_{\text{mass}}^{-adapt}$ achieves the same accuracy and inference cost as $\Delta_{\text{mass}}^{+adapt}$ with a budget of 2000 micro-clusters, despite using a median of 1750 context rows, compared with 1400 for the latter. More generally, at comparable inference costs, the adaptive configurations achieve higher accuracy. While the relative inference costs depend on the hardware and implementation, these results suggest that adaptation can improve the accuracy--cost trade-off without requiring larger contexts. Across both configurations, accuracy drops substantially when moving from all instances to the complete set of micro-clusters, then remains relatively stable at budgets of 1200--2000 before declining further at budgets of 400--800. This pattern suggests that moderate compression preserves most of the predictive information captured by the micro-cluster representation, whereas more aggressive compression discards information needed to represent the stream adequately. For $\Delta_{\text{mass}}^{+adapt}$, using all instances yields 84.5\% accuracy, with a median context size of 2650 rows and an inference cost of 120 ms/instance. Replacing all instances with micro-clusters reduces accuracy by only 0.8 percentage points while reducing context size by 37\% and inference time by 54\%. Imposing a budget of 2000 micro-clusters limits the accuracy loss to 1.0 percentage point relative to the full-instance baseline, while reducing context size by 60\% and inference time by 80\%.

\begin{figure*}[t]
\centering

\begin{minipage}[t]{0.73\textwidth}
\vspace{0pt}
\centering

\subcaptionbox{Total MC and core share\label{fig:vemos_a}}[0.32\linewidth][c]{%
    \includegraphics[width=\linewidth,height=0.19\textheight]{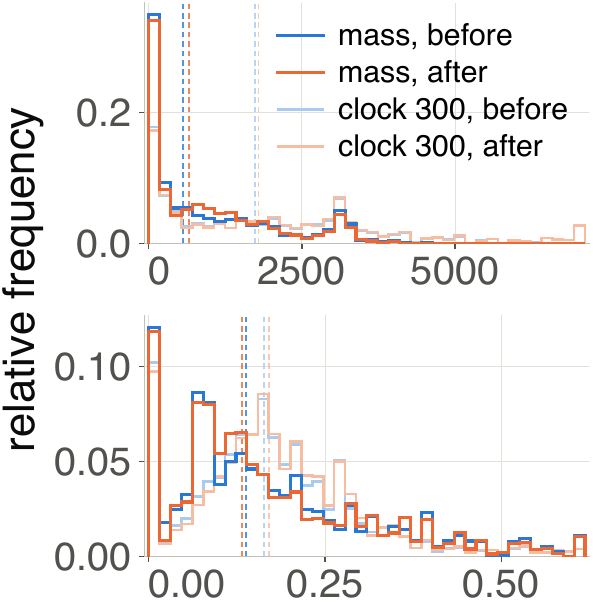}%
    
}\hfill
\subcaptionbox{Context size frequency for mass vs. clock 300 \label{fig:vemos_b}}[0.64\linewidth][c]{%
    \includegraphics[width=0.99\linewidth,height=0.19\textheight]{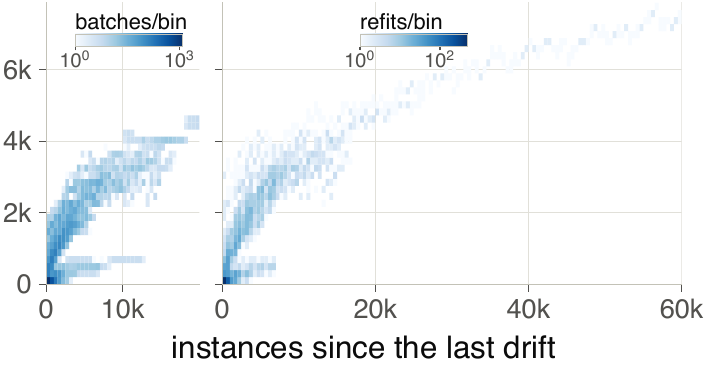}%
    
}

\vspace{2pt}

\subcaptionbox{Core role\label{fig:vemos_c}}[0.32\linewidth][c]{%
    \includegraphics[width=\linewidth,height=0.12\textheight]{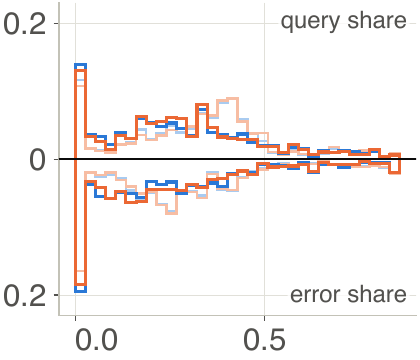}%
    
}\hfill
\subcaptionbox{Outlier role\label{fig:vemos_d}}[0.32\linewidth][c]{%
    \includegraphics[width=\linewidth,height=0.12\textheight]{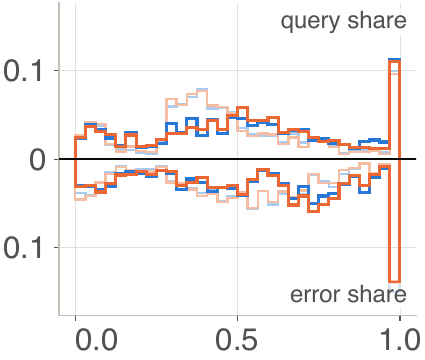}%
    
}\hfill
\subcaptionbox{Coverage signal\label{fig:vemos_e}}[0.32\linewidth][c]{%
    \includegraphics[width=\linewidth,height=0.12\textheight]{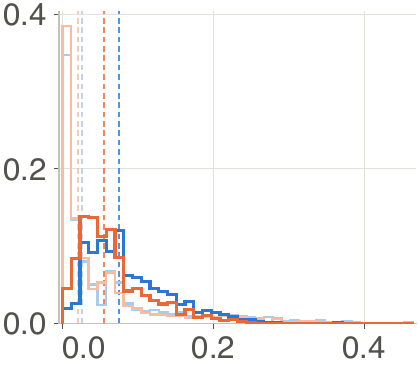}%
    
}

\vspace{2pt}

\subcaptionbox{Births/deaths\label{fig:vemos_f}}[0.32\linewidth][c]{%
    \includegraphics[width=\linewidth,height=0.12\textheight]{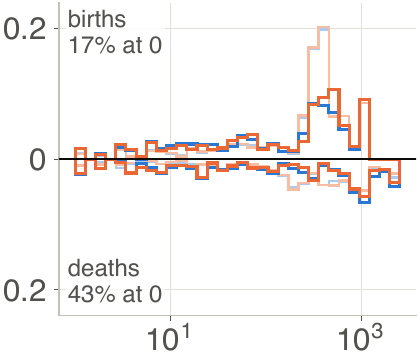}%
    
}\hfill
\subcaptionbox{Expansions/shrinkages\label{fig:vemos_g}}[0.32\linewidth][c]{%
    \includegraphics[width=\linewidth,height=0.12\textheight]{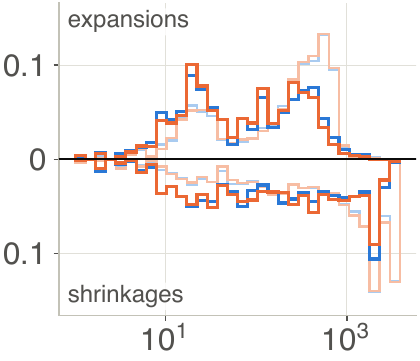}%
    
}\hfill
\subcaptionbox{Mass signal\label{fig:vemos_h}}[0.32\linewidth][c]{%
    \includegraphics[width=\linewidth,height=0.12\textheight]{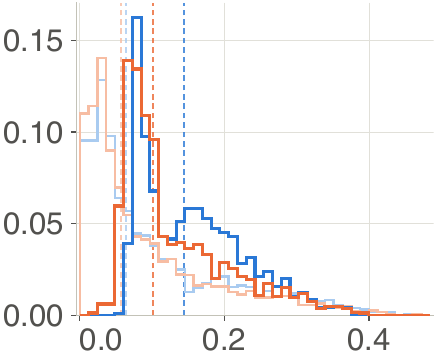 }%
    
}

\end{minipage}
\begin{minipage}[t]{0.26\textwidth}
\vspace{1pt}
\subcaptionbox{Median$\pm$IQR change at refits\label{fig:vemos_i}}[\linewidth][c]{%
    \includegraphics[width=0.95\linewidth,height=0.49\textheight]{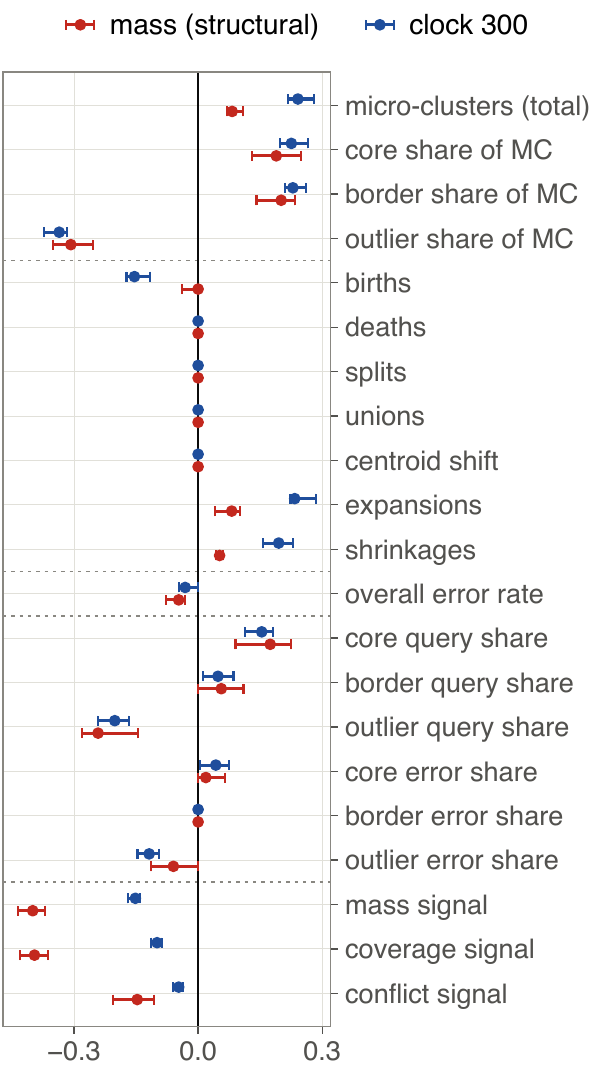}%
}
\end{minipage}
\caption{Structural changes and predictive outcomes under mass-discrepancy-triggered versus periodic refitting.}
\label{fig:vemos}
\end{figure*}

\textbf{Mass discrepancy signal allocates computation to more consequential structural changes (RQ3).} To characterize the relationship between structural evolution and predictive behaviour, SAIL's micro-cluster dynamics are analysed alongside prediction-query and error statistics. Three complementary views of structural evolution are considered: micro-cluster roles (core, border, and outlier), topological events (birth, death, split, and union), and geometric changes (expansion, shrinkage, and centroid movement). Full details are provided in Appendix~\ref{sec:structural-error-attribution}. Figure~\ref{fig:vemos} compares an extensive subset of these statistics under mass-discrepancy-triggered refitting and periodic refitting every 300 instances. Both policies incorporate performance-drift detection; however, periods affected by the associated restructuring are excluded because micro-cluster lineage cannot be reliably established. Comparing the two policies reveals differences in when structural changes occur over the stream. Because periodic refitting continues at a fixed rate, it allocates substantial computation to later stages of a longstanding concept, when the stream may provide less novel information. Consequently, as shown in Figure \ref{fig:vemos_a}, a larger share of measured transitions under periodic refitting involves core micro-clusters. Nevertheless, Figures~\ref{fig:vemos_c} and~\ref{fig:vemos_d} show that the shares of micro-cluster transitions, prediction queries, and errors across the three roles remain broadly comparable between the policies. Although this distribution of transitions across roles remains broadly comparable, Figure~\ref{fig:vemos_i} reveals adaptive refitting exhibits a wider spread in the expected transition, reflecting its tendency to concentrate adaptation around periods of more structural change.  

The nature of these transitions also differs. As shown in Figures~\ref{fig:vemos_f}, \ref{fig:vemos_g}, and~\ref{fig:vemos_i}, periodic refitting is associated with fewer births ($-0.15$), more shrinkages ($+0.19$) and more expansions ($+0.23$), and approximately unchanged numbers of deaths, splits, and unions ($0$). This pattern suggests that periodic refitting increasingly operates on an established micro-cluster structure, where evolution occurs through geometric adjustments rather than the creation of new clusters. Figure~\ref{fig:vemos_b} shows that periodic refitting sustains larger context sizes for longer, whereas adaptive refitting exhibits more dynamic changes in context size. In contrast, adaptive refitting is associated with smaller increases in shrinkages ($+0.05$ vs.~$+0.19$) and expansions ($+0.08$ vs.~$+0.23$), while births remain approximately unchanged, i.e., equally high, rather than decreasing ($0$ vs.~$-0.15$). These differences are consistent with adaptive refitting responding to structural changes while the representation is still developing, when new micro-clusters can capture emerging regions of the feature space, rather than repeatedly adjusting existing ones. Finally, adaptive refitting yields a larger drop in overall error rate ($-0.05$ vs.~$-0.03$), suggesting that \textit{mass discrepancy} helps direct adaptation towards periods of greater predictive benefit.

\textbf{Alternative sparse adaptation signals (RQ4).} Mass discrepancy tracks changes in the relative weights of micro-clusters, providing a general signal of structural redistribution. Two more targeted alternatives are considered to detect specific conditions that may warrant adaptation: class conflict, and geometric coverage gap. \textit{Conflict} measures changes in local class overlap. For each micro-cluster $\mu_k\in\mathcal{M}_c^{(t)}$, define its normalized inter-class margin as:
\begin{equation}
m_k=
\frac{1}{r_k}
\min_{c'\neq c}
\min_{\mu_j\in\mathcal{M}_{c'}^{(t)}}
\|\mathbf{z}_k-\mathbf{z}_j\|_2.
\end{equation}
The fraction of decayed mass in overlapping class regions is:
\begin{equation}
\Phi_{\mathrm{conflict}}(t)=
\frac{
\sum_c\sum_{\mu_k\in\mathcal{M}_c^{(t)}}
w_k^{(t)}\mathbb{1}[m_k<1]
}{
\sum_c\sum_{\mu_k\in\mathcal{M}_c^{(t)}}w_k^{(t)}
}.
\label{eq:conflict}
\end{equation}
\textit{Coverage} measures the fraction of stream mass lying beyond the geometric reach of the TFM context. Let $\mathcal{D}_{\mathrm{ctx}}^c$ denote its instances of class $c$. Then
\begin{equation}
\Phi_{\mathrm{coverage}}(t)=
\frac{
\sum_c\sum_{\mu_k\in\mathcal{M}_c^{(t)}}
w_k^{(t)}
\mathbb{1}\left[
\min_{\mathbf{x}\in\mathcal{D}_{\mathrm{ctx}}^c}
\|\mathbf{z}_k-\mathbf{x}\|_2>r_k
\right]
}{
\sum_c\sum_{\mu_k\in\mathcal{M}_c^{(t)}}w_k^{(t)}
}.
\label{eq:coverage}
\end{equation}
\begin{figure*}[t]
    \centering
    \begin{subfigure}{0.32\textwidth}
        \includegraphics[width=\linewidth]{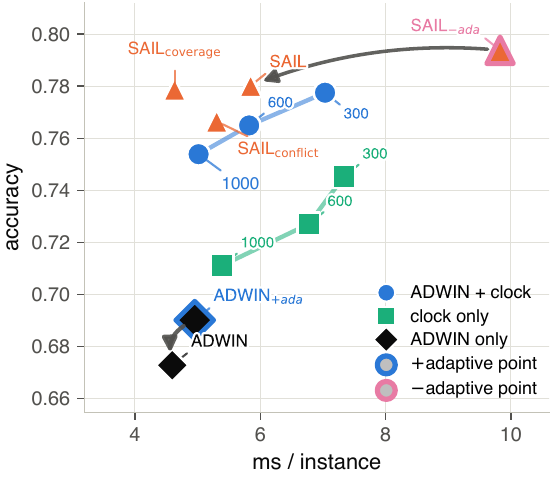}
        \caption{Accuracy vs. inference cost}
    \end{subfigure}
    \begin{subfigure}{0.32\textwidth}
        \includegraphics[width=\linewidth]{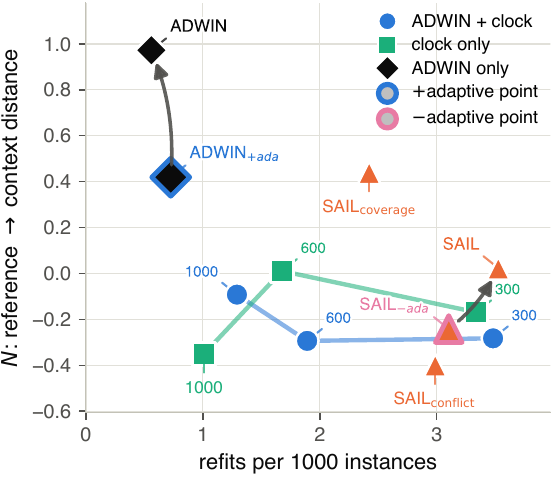}
        \caption{Coverage gap ($N$) vs. Refit frequency}
    \end{subfigure}
    \begin{subfigure}{0.32\textwidth}
        \includegraphics[width=\linewidth]{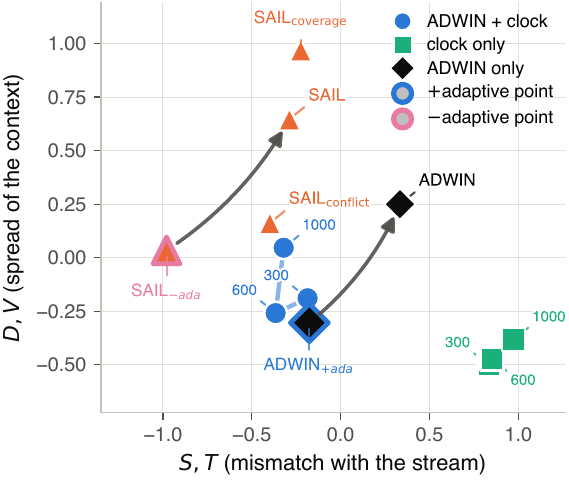}
        \caption{Diverse ($D,V$) vs. Representative ($S,T$)}
    \end{subfigure}
    \caption{Performance-based and mass discrepancy adaptation against alternative adaptive and fixed periodic policies.}
    \label{fig:detection1}
\end{figure*}
If a class is absent from the context, all its micro-cluster mass is considered uncovered. Each statistic defines a structural trigger based on its change, $\Delta_{\mathrm{conflict}}(t_0,t)$ or $\Delta_{\mathrm{coverage}}(t_0,t)$. Figures ~\ref{fig:vemos_e} and \ref{fig:vemos_h} provide an initial assessment of these signals by comparing mass-discrepancy-triggered and periodic refitting. Adaptive refitting yields larger reductions in mass discrepancy ($-0.40$ vs.~$-0.15$), coverage gap ($-0.40$ vs.~$-0.10$), and conflict ($-0.15$ vs.~$-0.05$). Although mass discrepancy directly tracks changes in relative micro-cluster weights, these results suggest that it can also guide adaptation towards improved context coverage and reduced class overlap.

To further assess these signals within SAIL's framework, Figure~\ref{fig:detection1} compares them as structural triggers against performance-based drift detection and periodic refitting, with and without performance-based detection. The context budget is fixed at 200 micro-clusters, with a clustering radius of $1.0$ and a cleanup interval of 1,000 instances. Performance-based detection is evaluated at ADWIN significance levels of $0.02$ and $0.002$. Structural triggers use a significance level of $0.002$ and a threshold of $0.10$. Periodic refitting is evaluated at intervals of 300, 600, and 1,000 instances. For reference, the $\Delta_{\mathrm{mass}}^{-adapt}$ configuration from Figure~\ref{fig:compression} is also included, which uses a clustering radius of $0.5$, a structural threshold of $0.25$, an ADWIN significance level of $0.02$, and a cleanup interval of 2,000 instances. Figure~\ref{fig:detection1} evaluates sparse adaptation along three complementary dimensions: predictive accuracy and inference cost, coverage gap to refitting frequency, as well as, temporal representativeness and geometric diversity. Feature-mean discrepancy ($S$), class-distribution discrepancy ($T$), and coverage gap ($N$), defined in Section~\ref{sec:preliminaries}, can also be used to characterize how well the context represents the incoming batch. However, they do not capture the internal geometric diversity of the context itself. We therefore introduce two additional descriptors: mean pairwise distance ($D$) and log-determinant covariance ($V$), which characterize spatial spread and multivariate dispersion, respectively. Given the context feature vectors $\{\mathbf{x}_i\}_{i=1}^{B}$, let $\Sigma_t$ denote their sample covariance matrix. Then:
\begin{equation}
D_t=
\frac{2}{B(B-1)}
\sum_{i<j}\|\mathbf{x}_i-\mathbf{x}_j\|_2,
\qquad
V_t=\log\det(\Sigma_t+\lambda I_d),
\end{equation}
where $\lambda>0$ is a fixed regularization parameter. $D_t$ measures average pairwise dispersion, whereas $V_t$ captures multivariate spread while accounting for feature correlations. Figure \ref{fig:detection1} shows error-based detection alone yields poor predictive accuracy and a large coverage gap, across both significance levels of $0.02$ and $0.002$. Although its contexts retain some geometric diversity ($D,V$), distributional discrepancies ($S,T$) remain poorly controlled. Periodic refitting reduces the coverage gap ($N$) and modestly improves accuracy, but does not consistently maintain geometric diversity or control these discrepancies. Combining periodic refitting with performance-based detection substantially reduces distributional discrepancies, while maintaining low coverage gaps, although geometric diversity remains less effectively controlled. Mass-discrepancy-triggered refitting combines these benefits with better preservation of geometric diversity, while achieving higher accuracy at lower inference cost. Importantly, the comparison of alternative structural signals further highlights the multidimensionality of context quality. The \textit{coverage} signal achieves the same predictive accuracy as \textit{mass discrepancy} at lower inference cost, owing to fewer refits, and preserves greater geometric diversity despite comparable distributional representativeness ($S,T$). However, its larger $N$ indicates that explicitly targeting coverage as a drift signal does not necessarily translate into better coverage of incoming instances. Conversely, \textit{conflict} achieves lower accuracy than both alternatives, comparable representativeness, smaller coverage, and lower geometric diversity. Importantly, beyond identifying \textit{mass discrepancy} as a balanced criterion, this ablation opens an avenue for further research.

\section{Conclusion}

This work introduced \textit{SAIL: Sparse Adaptation for In-Context Learning}, a framework for adapting TFMs to evolving data streams. SAIL separates routine inference from context reconstruction, reusing the key--value cache between adaptation events. Structural changes in an online micro-clustering representation trigger context reconstruction, while predictive-error monitoring guides updates to the micro-clustering structure itself. When adaptation is required, class-specific micro-clusters construct bounded contexts that prioritize recent observations while reducing redundancy.

The experimental results highlight three implications for streaming inference with TFMs. First, context size and adaptation frequency play distinct roles in the accuracy--cost trade-off: larger contexts increase inference cost and may worsen distributional discrepancies, while more frequent refitting has a stronger effect on predictive error. Context quality is also multidimensional, requiring a balance between distributional representativeness, geometric coverage, and diversity. Second, micro-cluster compression substantially reduces context size and inference cost with limited accuracy loss, although aggressive compression eventually discards predictive information. Third, mass-discrepancy-triggered adaptation improves the accuracy--cost trade-off while promoting reductions in coverage gaps and class conflict.

Future work follows two directions. First, SAIL's explicit characterization of predictive performance and computational trade-offs provides a basis for fairer comparisons between TFMs and conventional online learners, particularly fast Hoeffding-tree methods \cite{lourencco2026ihomer+,neves2025online}. Such comparisons should assess end-to-end latency and throughput under realistic edge workloads. Second, SAIL itself can be extended in several ways. Its reliance on Euclidean distances and micro-cluster geometry may limit performance in high-dimensional spaces, motivating subspace clustering methods that summarize both observations and relevant dimensions while updating only affected neighboring micro-clusters. Further work should also assess the robustness of structural signals across TFMs and streaming environments, develop principled strategies for setting structural thresholds and context budgets under computational constraints \cite{danilowski2026botta,lourencco2023automated}, and investigate model-aware structural signals that identify different forms of harmful drift in unlabeled settings \cite{neves2026pitfalls}.


\newpage

\printbibliography

\appendix

\begin{table*}[t]
\centering
\setlength{\tabcolsep}{1pt}
\renewcommand{\arraystretch}{0.79}
\caption{Accuracy, cost and dataset characteristics. Subscripts are the std. dev. across the evaluation batches of each run, and across the 12 streams in the \emph{Avg}. Bold marks the best value within each group. Robust $z$ data characteristics, for batch $B=600$.}
\label{tab:performance}
\begin{tabular}{l|c|cccc|ccccc|c|cccc}
\toprule
\multicolumn{2}{c}{}
& \multicolumn{4}{c}{\textbf{Refit trigger (200 micro-clusters)}}
& \multicolumn{5}{c}{\textbf{Budget sweep ($e{+}\Delta_{\text{mass}}^{-adapt}$)}}
& \multicolumn{5}{c}{\textbf{Data characteristics}} \\
\cmidrule{3-16}
\textbf{Data} & \textbf{Metric}
& \textbf{$e{+}\mathrm{ada}$}
& \textbf{$P_{300}$}
& \textbf{$e{+}P_{300}$}
& \textbf{$e{+}\Delta_{\mathrm{mass}}$}
& \textbf{200}
& \textbf{1200}
& \textbf{2000}
& \textbf{All MC}
& \textbf{All inst.}
& & \textbf{$E_{95}$} & \textbf{$P_{\kappa=2}$} & \textbf{$I_{\kappa=2}$} & \textbf{$C_{10}$} \\
\midrule
\multirow{3}{*}{AGR}
& {\footnotesize Acc} & 66.37$_{23.85}$ & 95.83$_{6.87}$ & 97.15$_{5.86}$ & \textbf{97.39$_{4.32}$} & 96.93$_{6.45}$ & 99.16$_{4.54}$ & 99.24$_{4.53}$ & 99.29$_{4.53}$ & \textbf{99.33$_{4.53}$} & S & 1.36 & 0.01 & -- & 0.30 \\
& {\footnotesize ms/inst} & \textbf{4.30$_{1.13}$} & 6.21$_{3.21}$ & 6.38$_{3.84}$ & 5.31$_{2.97}$ & \textbf{8.08$_{4.09}$} & 12.22$_{13.08}$ & 16.87$_{22.13}$ & 26.92$_{46.33}$ & 55.75$_{130.50}$ & T & 1.78 & 0.04 & 0.47 & 0.39 \\
& {\footnotesize ms/correct} & 6.48$_{1.70}$ & 6.48$_{3.35}$ & 6.57$_{3.95}$ & \textbf{5.45$_{3.05}$} & \textbf{8.33$_{4.22}$} & 12.32$_{13.19}$ & 17.00$_{22.30}$ & 27.12$_{46.66}$ & 56.13$_{131.38}$ & N & 1.56 & 0.02 & 0.00 & 0.32 \\

\multirow{3}{*}{COVER}
& {\footnotesize Acc} & 78.37$_{22.78}$ & \textbf{86.42$_{16.85}$} & 77.83$_{20.64}$ & 74.79$_{21.39}$ & 79.81$_{20.11}$ & 80.69$_{19.95}$ & 80.69$_{19.95}$ & 80.69$_{19.95}$ & \textbf{88.13$_{16.75}$} & S & 9.34 & 0.16 & -- & 0.67 \\
& {\footnotesize ms/inst} & 7.00$_{3.48}$ & 8.65$_{5.45}$ & 7.65$_{3.53}$ & \textbf{5.44$_{2.12}$} & \textbf{13.77$_{5.70}$} & 14.14$_{6.82}$ & 14.54$_{7.07}$ & 21.06$_{10.05}$ & 21.05$_{22.73}$ & T & 3.99 & 0.10 & 0.42 & 0.50 \\
& {\footnotesize ms/correct} & 8.93$_{4.44}$ & 10.01$_{6.31}$ & 9.83$_{4.53}$ & \textbf{7.27$_{2.83}$} & \textbf{17.25$_{7.15}$} & 17.52$_{8.45}$ & 18.02$_{8.76}$ & 26.10$_{12.46}$ & 23.89$_{25.79}$ & N & 18.48 & 0.17 & 1.55 & 0.85 \\

\multirow{3}{*}{ELEC}
& {\footnotesize Acc} & 74.92$_{15.34}$ & 76.50$_{14.72}$ & 75.54$_{15.10}$ & \textbf{76.75$_{14.18}$} & 78.26$_{13.98}$ & 78.69$_{14.12}$ & \textbf{78.70$_{14.11}$} & \textbf{78.70$_{14.11}$} & 78.09$_{13.85}$ & S & 2.33 & 0.09 & 0.99 & 0.36 \\
& {\footnotesize ms/inst} & 4.83$_{2.40}$ & 6.06$_{3.34}$ & 5.06$_{3.23}$ & \textbf{3.78$_{1.88}$} & 12.82$_{6.84}$ & \textbf{12.13$_{6.81}$} & 12.52$_{6.88}$ & 13.73$_{9.17}$ & 16.78$_{10.69}$ & T & 1.48 & 0.01 & -- & 0.26 \\
& {\footnotesize ms/correct} & 6.44$_{3.20}$ & 7.92$_{4.37}$ & 6.70$_{4.28}$ & \textbf{4.93$_{2.44}$} & 16.38$_{8.74}$ & \textbf{15.42$_{8.66}$} & 15.91$_{8.75}$ & 17.45$_{11.65}$ & 21.49$_{13.69}$ & N & 1.53 & 0.03 & 0.00 & 0.43 \\

\multirow{3}{*}{INS$_{abr.bal}$}
& {\footnotesize Acc} & 62.92$_{16.24}$ & 65.25$_{10.85}$ & 74.84$_{11.97}$ & \textbf{75.09$_{12.16}$} & 75.68$_{12.28}$ & 78.68$_{11.29}$ & 79.03$_{11.13}$ & 79.44$_{11.09}$ & \textbf{79.54$_{11.10}$} & S & 7.91 & 0.18 & 0.55 & 0.55 \\
& {\footnotesize ms/inst} & \textbf{5.10$_{1.70}$} & 7.60$_{4.77}$ & 7.84$_{4.55}$ & 7.95$_{5.08}$ & \textbf{9.22$_{5.50}$} & 16.52$_{18.53}$ & 17.58$_{23.68}$ & 37.22$_{75.56}$ & 19.51$_{51.43}$ & T & 14.11 & 0.18 & 0.32 & 0.58 \\
& {\footnotesize ms/correct} & \textbf{8.11$_{2.70}$} & 11.65$_{7.31}$ & 10.48$_{6.08}$ & 10.58$_{6.76}$ & \textbf{12.19$_{7.27}$} & 20.99$_{23.55}$ & 22.24$_{29.96}$ & 46.86$_{95.12}$ & 24.53$_{64.66}$ & N & 1.25 & 0.05 & 0.00 & 0.56 \\

\multirow{3}{*}{INS$_{grad.bal}$}
& {\footnotesize Acc} & 61.95$_{23.48}$ & 67.86$_{16.29}$ & 76.50$_{16.90}$ & \textbf{77.03$_{16.04}$} & 78.22$_{15.35}$ & \textbf{81.03$_{14.04}$} & 79.77$_{14.88}$ & 80.45$_{13.71}$ & 80.89$_{13.70}$ & S & 8.20 & 0.21 & 0.71 & 0.57 \\
& {\footnotesize ms/inst} & 4.81$_{1.86}$ & 7.28$_{4.31}$ & 7.52$_{4.41}$ & \textbf{4.50$_{2.88}$} & \textbf{10.72$_{6.42}$} & 15.66$_{16.79}$ & 19.43$_{25.11}$ & 31.00$_{57.13}$ & 44.22$_{128.12}$ & T & 8.48 & 0.23 & 0.56 & 0.56 \\
& {\footnotesize ms/correct} & 7.77$_{3.01}$ & 10.73$_{6.36}$ & 9.83$_{5.76}$ & \textbf{5.84$_{3.74}$} & \textbf{13.71$_{8.20}$} & 19.33$_{20.72}$ & 24.36$_{31.47}$ & 38.53$_{71.02}$ & 54.67$_{158.39}$ & N & 1.12 & 0.05 & 0.00 & 0.49 \\

\multirow{3}{*}{INS$_{inc.abr.bal}$}
& {\footnotesize Acc} & 67.61$_{16.55}$ & 62.94$_{13.77}$ & 74.56$_{14.98}$ & \textbf{75.12$_{14.42}$} & 75.53$_{14.13}$ & 78.18$_{13.23}$ & 78.59$_{13.04}$ & \textbf{79.01$_{12.72}$} & 78.75$_{12.95}$ & S & 3.19 & 0.14 & 0.35 & 0.46 \\
& {\footnotesize ms/inst} & \textbf{5.10$_{1.86}$} & 7.77$_{4.56}$ & 7.90$_{4.77}$ & 5.59$_{3.81}$ & \textbf{10.47$_{6.25}$} & 18.00$_{19.64}$ & 18.74$_{24.15}$ & 32.95$_{51.45}$ & 17.87$_{34.83}$ & T & 4.93 & 0.20 & 0.47 & 0.44 \\
& {\footnotesize ms/correct} & 7.55$_{2.76}$ & 12.35$_{7.24}$ & 10.60$_{6.39}$ & \textbf{7.44$_{5.07}$} & \textbf{13.86$_{8.27}$} & 23.03$_{25.12}$ & 23.85$_{30.72}$ & 41.70$_{65.12}$ & 22.70$_{44.23}$ & N & 1.69 & 0.04 & 0.40 & 0.56 \\

\multirow{3}{*}{INS$_{inc.rec.bal}$}
& {\footnotesize Acc} & 68.05$_{20.04}$ & 63.73$_{13.99}$ & 76.83$_{14.28}$ & \textbf{77.57$_{13.84}$} & 78.37$_{13.26}$ & 80.57$_{12.35}$ & 80.75$_{12.20}$ & \textbf{81.18$_{12.24}$} & 80.49$_{12.19}$ & S & 3.66 & 0.15 & 0.46 & 0.40 \\
& {\footnotesize ms/inst} & \textbf{4.46$_{1.60}$} & 7.91$_{4.85}$ & 8.08$_{5.22}$ & 7.96$_{5.40}$ & \textbf{9.94$_{6.01}$} & 17.84$_{19.43}$ & 15.73$_{20.82}$ & 33.97$_{54.50}$ & 28.38$_{60.44}$ & T & 5.13 & 0.17 & 0.45 & 0.43 \\
& {\footnotesize ms/correct} & \textbf{6.55$_{2.35}$} & 12.41$_{7.60}$ & 10.52$_{6.79}$ & 10.26$_{6.97}$ & \textbf{12.68$_{7.67}$} & 22.15$_{24.12}$ & 19.48$_{25.78}$ & 41.84$_{67.13}$ & 35.26$_{75.09}$ & N & 1.78 & 0.02 & 0.00 & 0.40 \\

\multirow{3}{*}{INS$_{inc.rec.imb}$}
& {\footnotesize Acc} & 67.82$_{8.63}$ & 71.38$_{9.15}$ & 74.71$_{7.17}$ & \textbf{75.13$_{7.20}$} & 76.01$_{6.93}$ & 78.50$_{6.85}$ & 78.84$_{6.81}$ & 79.41$_{6.71}$ & \textbf{79.59$_{7.04}$} & S & 3.29 & 0.07 & 0.32 & 0.48 \\
& {\footnotesize ms/inst} & \textbf{4.98$_{1.41}$} & 8.55$_{5.43}$ & 7.82$_{5.16}$ & 5.63$_{3.36}$ & \textbf{6.81$_{3.52}$} & 12.59$_{14.56}$ & 15.49$_{22.33}$ & 57.14$_{158.41}$ & 37.47$_{108.74}$ & T & 5.94 & 0.10 & 0.50 & 0.57 \\
& {\footnotesize ms/correct} & \textbf{7.34$_{2.07}$} & 11.98$_{7.60}$ & 10.47$_{6.90}$ & 7.49$_{4.47}$ & \textbf{8.95$_{4.63}$} & 16.04$_{18.54}$ & 19.65$_{28.32}$ & 71.96$_{199.49}$ & 47.08$_{136.62}$ & N & 0.78 & 0.00 & -- & 0.26 \\

\multirow{3}{*}{RBF}
& {\footnotesize Acc} & 76.75$_{8.92}$ & 73.21$_{7.91}$ & \textbf{80.61$_{4.80}$} & 79.94$_{6.82}$ & 81.21$_{6.13}$ & \textbf{82.48$_{6.47}$} & 82.32$_{6.71}$ & 82.30$_{6.60}$ & 82.45$_{6.50}$ & S & 1.20 & 0.02 & 0.00 & 0.27 \\
& {\footnotesize ms/inst} & 4.67$_{2.37}$ & 6.46$_{3.60}$ & 6.27$_{3.86}$ & \textbf{4.65$_{2.56}$} & \textbf{7.94$_{4.48}$} & 13.51$_{13.48}$ & 14.13$_{15.06}$ & 21.48$_{23.78}$ & 20.28$_{26.24}$ & T & 1.31 & 0.01 & 0.00 & 0.29 \\
& {\footnotesize ms/correct} & 6.08$_{3.09}$ & 8.83$_{4.92}$ & 7.78$_{4.79}$ & \textbf{5.82$_{3.20}$} & \textbf{9.78$_{5.51}$} & 16.38$_{16.34}$ & 17.16$_{18.30}$ & 26.10$_{28.89}$ & 24.60$_{31.83}$ & N & 1.29 & 0.00 & -- & 0.28 \\

\multirow{3}{*}{RIALTO}
& {\footnotesize Acc} & 47.57$_{30.45}$ & \textbf{64.51$_{27.54}$} & 57.69$_{27.47}$ & 60.94$_{25.89}$ & 66.17$_{26.80}$ & 65.88$_{26.64}$ & 65.88$_{26.64}$ & 65.88$_{26.64}$ & \textbf{68.55$_{26.79}$} & S & 1.58 & 0.02 & 0.00 & 0.28 \\
& {\footnotesize ms/inst} & \textbf{5.79$_{2.49}$} & 7.30$_{4.23}$ & 6.12$_{2.91}$ & 7.45$_{3.33}$ & 9.61$_{4.11}$ & 9.66$_{4.12}$ & 10.44$_{4.56}$ & 12.22$_{5.17}$ & \textbf{7.89$_{5.66}$} & T & -- & -- & -- & -- \\
& {\footnotesize ms/correct} & 12.17$_{5.23}$ & 11.32$_{6.56}$ & \textbf{10.60$_{5.05}$} & 12.22$_{5.47}$ & 14.52$_{6.21}$ & 14.67$_{6.25}$ & 15.84$_{6.93}$ & 18.55$_{7.85}$ & \textbf{11.51$_{8.26}$} & N & 2.44 & 0.11 & 0.36 & 0.34 \\

\multirow{3}{*}{TREE$_{recur}$}
& {\footnotesize Acc} & 72.27$_{11.62}$ & \textbf{82.97$_{6.98}$} & 82.66$_{7.07}$ & 82.74$_{7.00}$ & 83.07$_{7.31}$ & 91.82$_{5.45}$ & 93.29$_{5.24}$ & 94.68$_{5.50}$ & \textbf{95.69$_{5.46}$} & S & 1.76 & 0.03 & 0.00 & 0.25 \\
& {\footnotesize ms/inst} & \textbf{4.36$_{0.70}$} & 6.41$_{3.56}$ & 6.40$_{3.97}$ & 5.69$_{3.16}$ & \textbf{8.39$_{4.91}$} & 12.50$_{13.41}$ & 16.22$_{21.74}$ & 35.93$_{61.70}$ & 64.78$_{161.34}$ & T & 1.41 & 0.03 & 0.00 & 0.55 \\
& {\footnotesize ms/correct} & \textbf{6.03$_{0.97}$} & 7.72$_{4.29}$ & 7.74$_{4.80}$ & 6.88$_{3.82}$ & \textbf{10.10$_{5.91}$} & 13.61$_{14.61}$ & 17.39$_{23.30}$ & 37.94$_{65.17}$ & 67.70$_{168.61}$ & N & 0.81 & 0.00 & -- & 0.25 \\

\multirow{3}{*}{US$_{accident}$}
& {\footnotesize Acc} & 83.52$_{15.68}$ & 83.68$_{16.06}$ & \textbf{84.04$_{15.35}$} & 83.70$_{14.78}$ & 83.29$_{15.07}$ & 85.27$_{13.82}$ & 85.40$_{13.89}$ & 85.46$_{13.66}$ & \textbf{85.48$_{14.21}$} & S & 2.33 & 0.06 & 0.00 & 0.46 \\
& {\footnotesize ms/inst} & \textbf{4.05$_{1.38}$} & 7.79$_{4.49}$ & 7.36$_{4.87}$ & 6.21$_{4.04}$ & \textbf{10.16$_{5.78}$} & 17.58$_{18.61}$ & 16.60$_{20.65}$ & 25.77$_{35.33}$ & 31.44$_{60.98}$ & T & 2.70 & 0.12 & 0.79 & 0.38 \\
& {\footnotesize ms/correct} & \textbf{4.85$_{1.66}$} & 9.31$_{5.36}$ & 8.75$_{5.80}$ & 7.42$_{4.83}$ & \textbf{12.20$_{6.94}$} & 20.61$_{21.82}$ & 19.43$_{24.19}$ & 30.15$_{41.34}$ & 36.78$_{71.34}$ & N & 1.39 & 0.05 & 0.00 & 0.39 \\

\midrule
\multirow{4}{*}{Avg}
& {\footnotesize Acc} & 69.01$_{9.35}$ & 74.52$_{10.65}$ & 77.75$_{8.97}$ & \textbf{78.02$_{8.33}$} & 79.38$_{7.12}$ & 81.74$_{8.06}$ & 81.87$_{8.23}$ & 82.21$_{8.36}$ & \textbf{83.08$_{8.23}$} & \multicolumn{5}{l}{} \\
& {\footnotesize ms/inst} & \textbf{4.95$_{0.79}$} & 7.33$_{0.88}$ & 7.03$_{0.95}$ & 5.85$_{1.34}$ & \textbf{9.83$_{2.00}$} & 14.36$_{2.71}$ & 15.69$_{2.54}$ & 29.12$_{12.06}$ & 30.45$_{17.13}$ & \multicolumn{5}{l}{} \\
& {\footnotesize ms/correct} & \textbf{7.36$_{1.87}$} & 10.06$_{2.00}$ & 9.16$_{1.57}$ & 7.63$_{2.26}$ & \textbf{12.50$_{2.83}$} & 17.67$_{3.50}$ & 19.20$_{2.92}$ & 35.36$_{14.90}$ & 35.53$_{17.26}$ & \multicolumn{5}{l}{} \\
& {\footnotesize Rows} & 168$_{56}$ & 199$_{2}$ & 167$_{58}$ & \textbf{165$_{57}$} & \textbf{175$_{46}$} & 700$_{362}$ & 943$_{549}$ & 1370$_{1004}$ & 2068$_{1764}$ & \multicolumn{5}{l}{} \\
\multirow{4}{*}{Rank}
& {\footnotesize Acc} & 8.50 & 7.25 & 7.00 & 6.42 & 5.25 & 3.25 & 3.08 & 2.42 & \textbf{1.83} & \multicolumn{5}{l}{} \\
& {\footnotesize ms/inst} & \textbf{1.33} & 3.42 & 3.33 & 2.08 & 5.08 & 6.25 & 6.92 & 8.58 & 8.00 & \multicolumn{5}{l}{} \\
& {\footnotesize ms/correct} & \textbf{1.83} & 3.67 & 3.00 & \textbf{1.83} & 5.08 & 6.25 & 6.92 & 8.58 & 7.83 & \multicolumn{5}{l}{} \\
& {\footnotesize Rows} & \textbf{2.08} & 5.50 & 2.33 & 2.17 & 3.67 & 5.83 & 6.83 & 7.58 & 9.00 & \multicolumn{5}{l}{} \\
\bottomrule
\end{tabular}
\end{table*}

\begin{table}[t]
\centering
\caption{Sliding-window accuracy (mean $\pm$ SD across 12 streams) and mean refits per run.}
\label{tab:wp-accuracy}
\small
\setlength{\tabcolsep}{3pt}
\resizebox{\columnwidth}{!}{%
\begin{tabular}{@{}lrrrrrr@{}}
\toprule
 & \multicolumn{2}{c}{$P=500$}
 & \multicolumn{2}{c}{$P=1000$}
 & \multicolumn{2}{c}{$P=2000$} \\
\cmidrule(lr){2-3}
\cmidrule(lr){4-5}
\cmidrule(l){6-7}
$W$ & Accuracy & Refits & Accuracy & Refits & Accuracy & Refits \\
\midrule
100  & $0.730\pm0.103$ & 121 & $0.702\pm0.129$ & 61 & $0.667\pm0.155$ & 31 \\
200  & $0.767\pm0.106$ & 122 & $0.739\pm0.134$ & 62 & $0.696\pm0.161$ & 31 \\
500  & $0.798\pm0.106$ & 123 & $0.762\pm0.136$ & 62 & $0.716\pm0.165$ & 31 \\
1000 & $0.802\pm0.106$ & 124 & $0.769\pm0.136$ & 62 & $0.728\pm0.161$ & 31 \\
2000 & $0.799\pm0.109$ & 125 & $0.767\pm0.139$ & 63 & $0.728\pm0.160$ & 32 \\
5000 & $0.795\pm0.091$ & 126 & $0.768\pm0.112$ & 63 & $0.729\pm0.138$ & 32 \\
\bottomrule
\end{tabular}%
}
\end{table}

\section{Structural evolution and error attribution}
\label{sec:structural-error-attribution}

To investigate how structural evolution relates to predictive behavior, we analyze SAIL's micro-cluster dynamics alongside prediction queries, prequential errors, and adaptation events \cite{oliveira2010mec,risca2026collaborative}. We characterize structure through micro-cluster roles, changes in mass and centroid location, and topological transitions.

\textbf{Maintained statistics.}
SAIL maintains a class-conditional micro-cluster representation. Each micro-cluster $\mu_i$ has a persistent identifier, centroid $\mathbf{c}_i$, faded mass $w_i$, last-update timestamp $t_i$, and creation and last-absorption indices. Pairwise shared-density counters $s_{ij}$ record observations contributing to the neighborhoods of both micro-clusters. Masses and shared densities decay according to $2^{-\lambda(t-t')}$, where $t'$ denotes their last update. Structural snapshots are collected periodically, supporting structural analysis by tracking mass, centroid displacement, persistent identities, and density connectivity, without retaining member sets.

\textbf{Density-based roles.}
Following the shared-density principle \cite{cao2006density,hahsler2016clustering}, we construct an undirected graph $\mathcal{G}_c^{(t)}$ for each class, with micro-clusters as vertices. An edge of $\mu_i$ and $\mu_j$ is retained when
\begin{equation}
\frac{s_{ij}}{(w_i+w_j)/2} \geq \alpha,
\qquad \alpha=0.3.
\label{eq:shared-density-connectivity}
\end{equation}
Micro-clusters are classified as \emph{core} when they have at least one retained edge, \emph{border} when they have no retained edge but nonzero shared density with another micro-cluster, and \emph{outlier} otherwise.

\textbf{Structural transitions.}
For a micro-cluster surviving between consecutive snapshots, changes in faded mass and centroid location are quantified by
\begin{equation}
\Delta_w(i)=w_i(t+1)-w_i(t),
\qquad
\delta_i=
\|\mathbf{c}_i(t+1)-\mathbf{c}_i(t)\|_2.
\label{eq:structural-transitions}
\end{equation}
Positive and negative $\Delta_w$ indicate net mass growth and decay, respectively, while $\delta_i$ measures centroid displacement in the normalized feature space. Mass changes reflect the balance between newly absorbed observations and fading, rather than spatial expansion or contraction. Let $\mathcal{M}_t$ denote the micro-cluster identifiers at snapshot $t$. Births and deaths correspond to identifiers entering and leaving this set, while survival is identified by $\mathcal{M}_t\cap\mathcal{M}_{t+1}$. At the connected-component level, and restricted to micro-clusters that are core at both snapshots, a \emph{split} occurs when surviving members of one component belong to multiple components at the next snapshot; a \emph{union} occurs when surviving members of multiple components become part of the same component. These persistent identifiers therefore allow structural transitions to be characterized without observation-level matching or correspondence thresholds.

\textbf{Error attribution.} Let $q_i$ and $e_i$ denote the queries and prequential errors attributed to micro-cluster $\mu_i$ within a snapshot interval. For the set $\mathcal{S}_\rho$ of micro-clusters assigned role $\rho$, we compute the query share, and error share:
\begin{equation}
u_\rho =
\frac{\sum_{i\in\mathcal{S}_\rho}q_i}{\sum_i q_i},
\quad
v_\rho =
\frac{\sum_{i\in\mathcal{S}_\rho}e_i}{\sum_i e_i}
\label{eq:role-attribution}
\end{equation}
Undefined ratios are excluded when their denominators vanish. Comparing $u_\rho$ and $v_\rho$ reveals whether a structural role accounts for a disproportionate share of prediction errors.

\textbf{Event-centered comparison.}
Each quantity is compared across an adaptation event. For a refit at stream index $\tau$ and a horizon of $h$ instances, every measure $z$ is averaged over snapshots in $(\tau-h,\tau]$ and $(\tau,\tau+h]$, and the paired change
$\Delta z(\tau)=z^{\mathrm{post}}-z^{\mathrm{pre}}$
is recorded. To account for differences in scale, each $\Delta z$ is divided by the interquartile range of its distribution pooled over the compared policies. Changes are summarized by their median and a $95\%$ percentile-bootstrap interval based on $2000$ resamples.

\textbf{Experimental configuration.}
The micro-cluster radius is scaled with dimensionality as $r=2\sqrt{d/33}$ in the normalized feature space, using $d=33$ and $r=2$ as the reference configuration. We use fading factor $\lambda=10^{-4}$, cleanup interval $1000$, minimum retained mass $w_{\min}=1$, maximum $M_{\max}=2000$ micro-clusters per class, and shared-density threshold $\alpha=0.3$. The analysis compares fixed-clock adaptation at $300$ instances with the structural mass-trigger policy, both retaining performance-drift detection. Structural continuity is evaluated separately across memory resets, since rebuilding invalidates persistent micro-cluster lineage.

\section{Results}
\label{sec:results}

Performance was assessed using prequential \emph{test-then-train} evaluation across 12 streams, described in Table \ref{tab:streams}. Table \ref{tab:performance} highlights some of SAIL variants from Figures \ref{fig:compression} and \ref{fig:detection1}, using contexts compressed to 200 recent observations, 200 micro-clusters, or all micro-clusters, with adaptation triggered by error-rate drift ($e$), periodic intervals of 300, or mass drift at varying sensitivities.  It also reports the 4 temporal characteristics of coverage gap, feature-mean, and class-distribution discrepancy across selected streams: relative excursion magnitude, elevated-state prevalence, temporal intermittency, and relative change concentration. Table \ref{tab:wp-accuracy} shows accuracy for fixed policies. Table \ref{tab:profiles_remaining} summarizes the other 68 streams considered. 

\begin{figure}[t]
    \centering
    \begin{subfigure}{0.32\linewidth}
        \includegraphics[width=\linewidth]{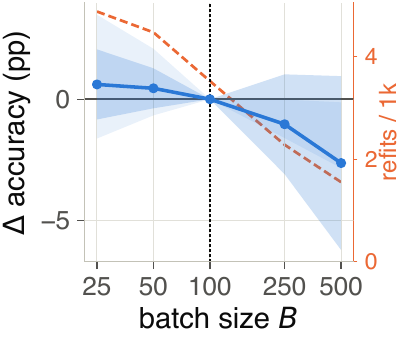}
        \caption{Predict batch size}
    \end{subfigure}
    \begin{subfigure}{0.32\linewidth}
        \includegraphics[width=\linewidth]{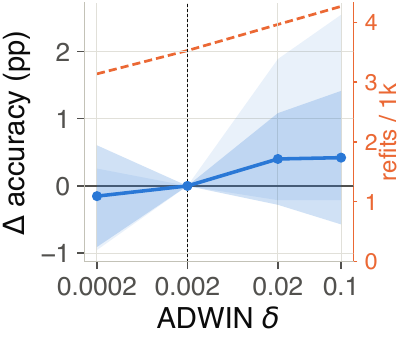}
        \caption{Performance drift}
    \end{subfigure}
    \begin{subfigure}{0.32\linewidth}
        \includegraphics[width=\linewidth]{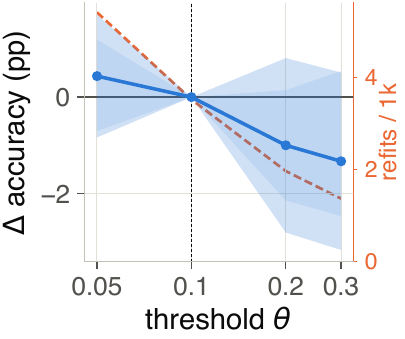}
        \caption{Structural drift}
    \end{subfigure}
    \caption{Sparse adaptation hyperparameters.}
    \label{fig:sparseadaptation}
\end{figure}

\begin{figure}[t]
    \centering
    \begin{subfigure}{0.32\linewidth}
        \includegraphics[width=\linewidth]{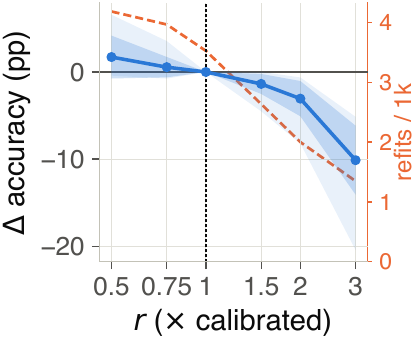}
        \caption{Radius}
    \end{subfigure}
    \begin{subfigure}{0.32\linewidth}
        \includegraphics[width=\linewidth]{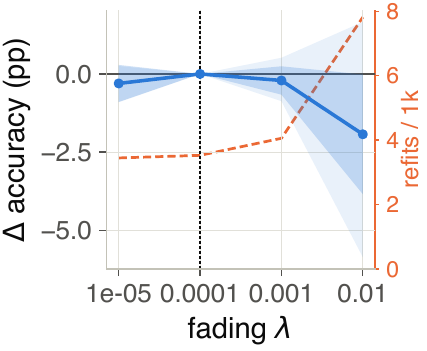}
        \caption{Fading weights}
    \end{subfigure}
    \begin{subfigure}{0.32\linewidth}
        \includegraphics[width=\linewidth]{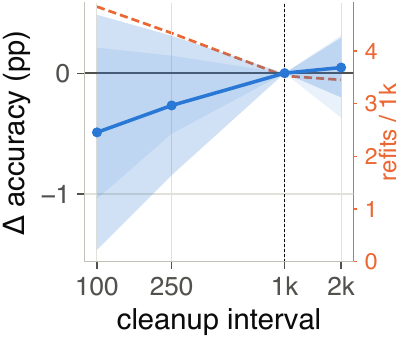}
        \caption{Cleanup interval}
    \end{subfigure}
    \caption{Micro-clustering hyperparameters.}
    \label{fig:microclustering}
\end{figure}

\textbf{Hyperparameters.} The sensitivity of SAIL to its hyperparameters is evaluated across alternative configurations and random seeds. Figures \ref{fig:sparseadaptation} and \ref{fig:microclustering} report the change in predictive accuracy relative to a reference configuration (blue), alongside the corresponding adaptation frequency (orange). For batch size, we consider \(B \in \{25,50,100,250,500\}\); structural-drift sensitivity, \(\theta \in \{0.05,0.1,0.2,0.3\}\); micro-cluster radius, \(r \in \{0.5,0.75,1.5,2,3\}\), expressed as a multiple of its calibrated value; fading factor, \(\lambda \in \{10^{-5},10^{-4},10^{-3},10^{-2}\}\); cleanup interval, \(C \in \{100,250,1000,2000\}\); and performance-drift sensitivity, \(\delta \in \{0.0002,0.002,0.02,0.1\}\). Accuracy generally declines with larger batches, structural-drift thresholds, and micro-cluster radii, with pronounced degradation at extreme values. A similar decline occurs at $(\lambda=10^{-2})$. In contrast, accuracy remains comparatively stable across performance-drift sensitivities and cleanup intervals. Adaptation frequency varies substantially across configurations.

\newpage

\begin{figure}[t]
    \centering
    \begin{subfigure}{0.25\linewidth}
        \includegraphics[width=\linewidth]{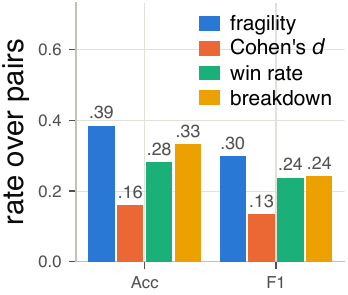}
        \caption{Fragility rate}
        \label{fig:robust_0}
    \end{subfigure}
    \begin{subfigure}{0.24\linewidth}
        \includegraphics[width=\linewidth]{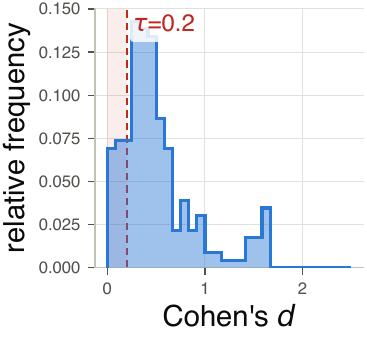}
        \caption{Cohen's $d$}
        \label{fig:robust_1}
    \end{subfigure}
    \begin{subfigure}{0.24\linewidth}
        \includegraphics[width=\linewidth]{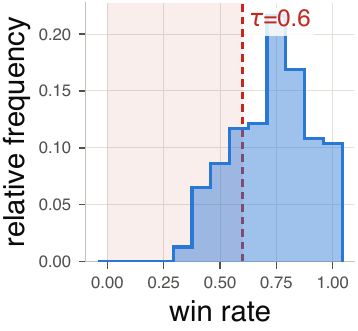}
        \caption{Win rate}
        \label{fig:robust_2}
    \end{subfigure}
    \begin{subfigure}{0.24\linewidth}
        \includegraphics[width=\linewidth]{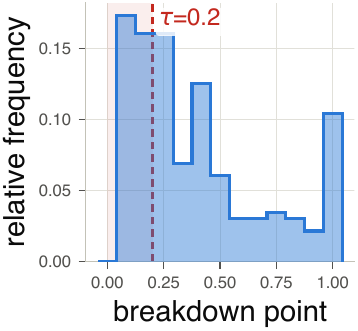}
        \caption{Break Point}
        \label{fig:robust_3}
    \end{subfigure}
    \caption{Robustness across all hyperparameter choices.}
    \label{fig:robust}
\end{figure}

\textbf{Robustness.} Table \ref{tab:performance} ranks a subset of SAIL variants by their mean performance, $\bar{s}_m=\frac{1}{k}\sum{d\in\mathcal{D}}s_m(d)$, with $A$ outperforming $B$ if $\bar{s}_A>\bar{s}_B$, given streams $\mathcal{D}={d_1,\ldots,d_k}$. Beyond average differences, for each pair of all configurations, we analyze the per-dataset difference $\delta_d(A,B)=s_A(d)-s_B(d)$ through four complementary criteria: Cohen's $d$, Win rate ($WR$), Breakdown point ($BP$), and Fragility ($F$). For ablation studies, $A$ and $B$ represent all ablations of the six hyperparameter choices, with Fig. \ref{fig:robust} summarizing the overall proportion of comparisons failing each diagnostic criterion. Cohen's $d$ assesses whether improvements are meaningful relative to cross-dataset variability through the standardized effect size $d(A,B)=\bar{\delta}/s_\delta$. $WR$ measures how frequently \(A\) outperforms \(B\) across datasets, 
\(WR(A,B)=\frac{1}{k}\sum_{d\in\mathcal{D}}\mathbb{1}[\delta_d(A,B)>0]\).
$BP$ quantifies ranking robustness under dataset removal, 
\(BP(A,B)=\frac{1}{k}\min_{S\subset\mathcal{D}}\{|S|:\bar{s}^{(-S)}_A\leq\bar{s}^{(-S)}_B\}\).
$F$ measures the fraction of comparisons failing at least one the three criterion,
\(F=\frac{1}{n}\sum_{(A,B)}\mathbb{1}[WR\leq\tau_w\vee d\leq\tau_d\vee BP\leq\tau_b]\). Across ablations, Figure \ref{fig:robust_0} shows only 16\%, 28\%, and 33\% of comparisons fail the criterion in accuracy, and 13\%, 24\%, and 24\% in F1-score. The resulting fragility rate is 39\%, and 30\%, respectively. Fig. \ref{fig:robust_1}, \ref{fig:robust_2}, and \ref{fig:robust_3} detail the respective distributions of Cohen's $d$, Win rate, Breakdown point for accuracy.

\begin{table}[t]
\centering
\caption{List of 68 benchmarks considered, but not selected.}
\label{tab:profiles_remaining}
\setlength{\tabcolsep}{1pt}
\footnotesize
\renewcommand{\arraystretch}{0.8}
\begin{tabular}{lrrrrrrrrr}
\toprule
Stream & $n$ & $d$ & $C$ & $S$ & $T$ & $N$ & $Q$ & transf. & cd \\
\midrule
outdoor\textsuperscript{4} & 4\,000 & 21 & 40 & .467 & .806 & .377 & .550 & .494 & .060 \\
readmission\textsuperscript{8} & 99\,493 & 47 & 2 & .842 & .672 & .491 & .668 & .578 & .014 \\
in\_vehicle\_coupon\textsuperscript{4} & 12\,684 & 24 & 2 & .327 & .653 & .327 & .435 & .597 & .022 \\
ins\_inc\_bal\textsuperscript{1,3} & 57\,018 & 33 & 6 & .588 & .434 & .252 & .424 & .625 & .000 \\
airlines\textsuperscript{1} & 100\,000 & 7 & 2 & .725 & .547 & .719 & .664 & .655 & .000 \\
ecom\_offers\textsuperscript{4,7} & 50\,000 & 119 & 2 & .623 & .700 & .666 & .663 & .660 & .031 \\
ecommerce\_shipping\textsuperscript{4} & 10\,999 & 10 & 2 & .595 & .275 & .402 & .424 & .666 & .000 \\
bioresponse\textsuperscript{4} & 3\,751 & 1776 & 2 & .448 & .791 & .433 & .557 & .713 & .000 \\
taxi\textsuperscript{2} & 42\,798 & 7 & 2 & .777 & .395 & .812 & .661 & .713 & .000 \\
covertype\_beyondarena\textsuperscript{4} & 100\,000 & 13 & 3 & .211 & .289 & .320 & .273 & .733 & .000 \\
predict\_students\_dropout\_and\_.\textsuperscript{4} & 4\,424 & 36 & 3 & .756 & .525 & .420 & .567 & .740 & .000 \\
labour\_inspection\_compliance\textsuperscript{4} & 63\,634 & 376 & 2 & .459 & .342 & .473 & .425 & .742 & .000 \\
led\_drift\textsuperscript{5} & 50\,000 & 24 & 10 & .441 & .362 & .423 & .409 & .742 & .000 \\
acs\_mobility\textsuperscript{8} & 50\,000 & 21 & 2 & .392 & .338 & .380 & .370 & .745 & .000 \\
ins\_inc\_imbal\textsuperscript{1,3} & 452\,044 & 33 & 6 & .438 & .612 & .270 & .440 & .750 & .021 \\
ins\_abrupt\_imbal\textsuperscript{1,3} & 355\,275 & 33 & 6 & .525 & .684 & .347 & .519 & .752 & .034 \\
acs\_pubcov\textsuperscript{8} & 50\,000 & 19 & 2 & .625 & .527 & .555 & .569 & .755 & .000 \\
ins\_inc\_abrupt\_imbal\textsuperscript{1,3} & 452\,044 & 33 & 6 & .716 & .784 & .353 & .618 & .761 & .040 \\
poker\textsuperscript{1} & 100\,000 & 10 & 8 & .516 & .456 & .750 & .574 & .761 & .164 \\
hotel\_booking\_demand\textsuperscript{4} & 81\,418 & 30 & 2 & .344 & .283 & .637 & .421 & .763 & .018 \\
acs\_income\textsuperscript{8} & 50\,000 & 10 & 2 & .439 & .242 & .527 & .403 & .764 & .000 \\
hr\_analytics\textsuperscript{4} & 19\,158 & 12 & 2 & .270 & .300 & .588 & .386 & .772 & .000 \\
noaa\textsuperscript{1,3} & 18\,159 & 8 & 2 & .372 & .306 & .662 & .447 & .775 & .000 \\
wids\_diabetes\_mellitus\textsuperscript{4} & 100\,000 & 181 & 2 & .263 & .173 & .383 & .273 & .788 & .000 \\
acquire\_valued\_shoppers\textsuperscript{4} & 100\,000 & 110 & 2 & .803 & .633 & .822 & .753 & .797 & .008 \\
jm1\textsuperscript{4} & 10\,885 & 21 & 2 & .775 & .253 & .572 & .533 & .805 & .018 \\
credit\_card\_clients\_default\textsuperscript{4} & 30\,000 & 23 & 2 & .359 & .256 & .528 & .381 & .806 & .000 \\
sepsis\_survival\_minimal\_clini.\textsuperscript{4} & 100\,000 & 3 & 2 & .339 & .550 & .453 & .447 & .810 & .004 \\
acs\_foodstamps\textsuperscript{8} & 50\,000 & 20 & 2 & .581 & .475 & .483 & .513 & .813 & .000 \\
ins\_gradual\_imbal\textsuperscript{1,3} & 143\,323 & 33 & 6 & .534 & .652 & .311 & .499 & .814 & .032 \\
kickstarter\textsuperscript{4} & 100\,000 & 14 & 2 & .594 & .456 & .539 & .530 & .834 & .017 \\
bank\_customer\_churn\textsuperscript{4} & 10\,000 & 10 & 2 & .405 & .331 & .452 & .396 & .835 & .000 \\
homesite\_quote\_conversion\textsuperscript{4} & 100\,000 & 295 & 2 & .286 & .387 & .436 & .370 & .838 & .007 \\
homeq\_default\textsuperscript{4} & 5\,708 & 12 & 2 & .723 & .263 & .367 & .451 & .841 & .000 \\
sea\textsuperscript{5} & 10\,000 & 3 & 2 & .355 & .614 & .470 & .480 & .848 & .000 \\
homesite\textsuperscript{7} & 20\,000 & 299 & 2 & .380 & .550 & .534 & .488 & .863 & .027 \\
kick\textsuperscript{4} & 72\,983 & 31 & 2 & .338 & .333 & .363 & .344 & .871 & .000 \\
bank\_marketing\textsuperscript{4} & 45\,211 & 13 & 2 & .266 & .470 & .494 & .410 & .883 & .000 \\
online\_shoppers\_purchasing\_in.\textsuperscript{4} & 12\,330 & 17 & 2 & .330 & .547 & .311 & .396 & .884 & .000 \\
ghanas\_indigenous\_intel\textsuperscript{4} & 10\,928 & 9 & 4 & .683 & .631 & .678 & .664 & .887 & .059 \\
customer\_satisfaction\_in\_airl.\textsuperscript{4} & 100\,000 & 21 & 2 & .461 & .409 & .433 & .434 & .887 & .006 \\
churn\textsuperscript{4} & 5\,000 & 19 & 2 & .358 & .412 & .592 & .454 & .892 & .000 \\
naticusdroid\_android\_permissi.\textsuperscript{4} & 7\,491 & 85 & 2 & .277 & .272 & .680 & .409 & .892 & .000 \\
santander\_customer\_transaction\textsuperscript{4} & 100\,000 & 600 & 2 & .375 & .469 & .280 & .374 & .892 & .000 \\
musk\textsuperscript{4} & 6\,598 & 166 & 2 & .575 & .822 & .395 & .597 & .903 & .194 \\
diabetes\_130\_us\textsuperscript{4} & 71\,518 & 44 & 2 & .284 & .550 & .628 & .488 & .912 & .000 \\
hyperplane\textsuperscript{5} & 50\,000 & 10 & 2 & .409 & .444 & .377 & .410 & .913 & .015 \\
home\_credit\_default\_risk\textsuperscript{4} & 100\,000 & 504 & 2 & .320 & .334 & .452 & .369 & .913 & .000 \\
polish\_companies\_bankruptcy\textsuperscript{4} & 5\,790 & 64 & 2 & .683 & .509 & .386 & .526 & .927 & .000 \\
give\_me\_some\_credit\textsuperscript{4} & 100\,000 & 10 & 2 & .398 & .311 & .744 & .484 & .932 & .000 \\
coil\_2000\textsuperscript{4} & 9\,822 & 85 & 2 & .395 & .739 & .470 & .535 & .936 & .000 \\
amazon\_employee\_access\textsuperscript{4} & 32\,769 & 9 & 2 & .361 & .425 & .681 & .489 & .943 & .000 \\
splice\textsuperscript{4} & 3\,190 & 60 & 3 & .495 & .441 & .875 & .604 & .945 & .000 \\
pva\_revenue\_kddcup98\textsuperscript{4} & 100\,000 & 477 & 2 & .448 & .431 & .439 & .440 & .945 & .000 \\
porto\_seguro\textsuperscript{4} & 100\,000 & 37 & 2 & .380 & .227 & .406 & .337 & .961 & .000 \\
santander\_customer\_satisfacti.\textsuperscript{4} & 71\,080 & 307 & 2 & .228 & .444 & .525 & .399 & .962 & .000 \\
sdss\_17\textsuperscript{4} & 78\,053 & 11 & 3 & .503 & .308 & .397 & .403 & .963 & .004 \\
hiva\_agnostic\textsuperscript{4} & 3\,845 & 1518 & 2 & .414 & .341 & .423 & .393 & .965 & .000 \\
emscad\textsuperscript{4} & 17\,460 & 17 & 2 & .689 & .645 & .594 & .643 & .967 & .102 \\
taiwanese\_bankruptcy\textsuperscript{4} & 6\,819 & 92 & 2 & .545 & .291 & .778 & .538 & .967 & .000 \\
acs\_unemployment\textsuperscript{8} & 50\,000 & 17 & 2 & .713 & .500 & .411 & .541 & .972 & .004 \\
homecredit\textsuperscript{7} & 10\,000 & 696 & 2 & .691 & .397 & .433 & .507 & .973 & .008 \\
ieee\_fraud\_detection\textsuperscript{4} & 100\,000 & 434 & 2 & .559 & .369 & .619 & .516 & .973 & .006 \\
kdd\_cup\_09\_appetency\textsuperscript{4} & 50\,000 & 212 & 2 & .427 & .397 & .550 & .458 & .982 & .006 \\
aps\_failure\textsuperscript{4} & 76\,000 & 170 & 2 & .463 & .320 & .475 & .419 & .982 & .000 \\
thyroid\_discordant\textsuperscript{4} & 3\,711 & 26 & 2 & .442 & .880 & .461 & .594 & .985 & .007 \\
otto\_group\textsuperscript{4} & 61\,878 & 93 & 9 & .734 & .927 & .477 & .713 & 1.00 & .000 \\
kddcup99\textsuperscript{4} & 100\,00 & 41 & 6 & .812 & .927 & .802 & .847 & 1.00 & .000 \\
\bottomrule
\end{tabular}
\vspace{2pt}
\begin{minipage}{\linewidth}
\footnotesize
\textsuperscript{1} USP \cite{souza2020challenges}
\textsuperscript{2} WhyShift \cite{liu2023need}
\textsuperscript{3} OEBench \cite{diao2023oebench}
\textsuperscript{4} BeyondArena \cite{purucker2026beyond}
\textsuperscript{5} Synth
\textsuperscript{6} TDG
\textsuperscript{7} TabRED \cite{rubachev2024tabred}
\textsuperscript{8} TableShift \cite{gardner2023benchmarking}
\end{minipage}
\end{table}

\section{Reproducibility}

All experiments use a single-pass prequential (test-then-train)  evaluation in which predictions are made before labels are revealed. The evaluation was conducted on an Apple M4 (Mac16,12) CPU host with 10 cores (4 performance and 6 efficiency cores) and 16 GiB of unified memory. The system ran macOS 27.0.1 (build 26A434) with Darwin kernel 27.0.0. The software environment used Python 3.13.7, PyTorch 2.14.1 with four CPU threads per process, NumPy 2.3.5, and SciPy 1.17.0. The tabular foundation model backbone was CausiLo 1.0.3 \cite{cho2026causilo}, configured with eight estimators and KV-cache reuse enabled. The underlying micro-clustering machinery was implemented using River 0.23.0. Experiments ran two concurrent processes, each using four threads, for a total of eight threads across the 10-core CPU. The code is available at [link after acceptance], together with dataset preparation and evaluation scripts, complete experimental configurations, and per-window metrics and histories.

\newpage

\end{document}